\documentclass{article}

\usepackage{graphicx}
\usepackage{cite}
\usepackage{amsmath}

\newcommand{\widefigwidth}{5.0in}
\newcommand{\figwidth}{3.5in}
\newcommand{\mfigwidth}{3.0in}
\newcommand{\smallfigwidth}{2in}

\newcommand{\eq}[1]{Eq.~(\ref{eq.#1})} 
\newcommand{\fig}[1]{Fig.~\ref{fig.#1}}

\newcommand{\tbl}[1]{Table~\ref{table.#1}}

\newcommand{\sect}[1]{Section~\ref{sect.#1}}
\newcommand{\sectA}[1]{Appendix~\ref{sect.#1}}
\newcommand{\sectlabel}[1]{\label{sect.#1}}
\newcommand{\eqlabel}[1]{\label{eq.#1}}
\newcommand{\figlabel}[1]{\label{fig.#1}}
\newcommand{\tbllabel}[1]{\label{table.#1}}

\newcommand{\flux}{\ensuremath{\mathcal{F}}}
\newcommand{\soundSpeed}{\ensuremath{c}}
\newcommand{\pIncident}{\ensuremath{p_{\textnormal{inc}}}} 
\newcommand{\pScattered}{\ensuremath{p_{\textnormal{s}}}} 
\newcommand{\vIncident}{\ensuremath{v_{\textnormal{inc}}}} 
\newcommand{\fluxIncident}{\ensuremath{\flux_{\textnormal{inc}}}} 
\newcommand{\fluxScattered}{\ensuremath{\flux_{\textnormal{s}}}} 

\newcommand{\density}{\ensuremath{\rho}}
\newcommand{\viscosity}{\ensuremath{\eta}}
\newcommand{\viscosityBulk}{\ensuremath{\eta_{\textnormal{bulk}}}}
\newcommand{\Tfluid}{\ensuremath{T_{\textnormal{fluid}}}}
\newcommand{\kThermal}{\ensuremath{k_{\textnormal{thermal}}}}
\newcommand{\heatCapacity}{\ensuremath{C_{\textnormal{p}}}} 
\newcommand{\heatCapacityRatio}{\ensuremath{\gamma}} 
\newcommand{\absorptionTissue}{\ensuremath{\alpha_{\textnormal{tissue}}}} 

\newcommand{\bodyVolume}{ \ensuremath{V_{\textnormal{body}}} }

\newcommand{\rRobot}{\ensuremath{r_{\textnormal{robot}}}} 
\newcommand{\numberDensityRobot}{\ensuremath{\nu_{\textnormal{robot}}}} 
\newcommand{\attenuationRobot}{\ensuremath{\alpha_{\textnormal{robot}}}} 

\newcommand{\Fdrag}{\ensuremath{F_{\textnormal{drag}}}}
\newcommand{\kTotal}{\ensuremath{k_{\textnormal{total}}}}
\newcommand{\kFriction}{\ensuremath{k_{\textnormal{f}}}}
\newcommand{\kViscous}{\ensuremath{k_{\textnormal{viscous}}}}
\newcommand{\kSliding}{\ensuremath{k_{\textnormal{sliding}}}}
\newcommand{\kLoad}{\ensuremath{k_{\textnormal{load}}}}
\newcommand{\kLoadRatio}{\ensuremath{k_{\textnormal{ratio}}}}
\newcommand{\ASliding}{\ensuremath{A_{\textnormal{sliding}}}} 
\newcommand{\ASlidingPiston}{\ensuremath{\ASliding^{\textnormal{piston}}}}

\newcommand{\Fspring}{\ensuremath{F_{\textnormal{spring}}}}
\newcommand{\Pambient}{\ensuremath{p_{\textnormal{ambient}}}}
\newcommand{\Kspring}{\ensuremath{K_{\textnormal{spring}}}}
\newcommand{\Lspring}{\ensuremath{L_{\textnormal{spring}}}}
\newcommand{\hspring}{\ensuremath{h_{\textnormal{spring}}}}
\newcommand{\FdragSpring}{\ensuremath{\Fdrag^{\textnormal{spring}}}}

\newcommand{\Hamaker}{\ensuremath{H_{\textnormal{vdW}}}} 
\newcommand{\PvdW}{\ensuremath{p_{\textnormal{vdW}}}} 

\newcommand{\fPistons}{\ensuremath{f_\textnormal{surface}}}

\newcommand{\PTotal}{\ensuremath{P_{\textnormal{total}}}}
\newcommand{\PLoad}{\ensuremath{P_{\textnormal{load}}}}

\newcommand{\meter}{\mbox{m}}
\newcommand{\centimeter}{\mbox{cm}}
\newcommand{\millimeter}{\mbox{mm}}
\newcommand{\micron}{\mbox{$\mu$m}}
\newcommand{\nanometer}{\mbox{nm}}
\newcommand{\liter}{\mbox{L}}
\newcommand{\second}{\mbox{s}}

\newcommand{\microsecond}{\mbox{$\mu$s}}
\newcommand{\kilogram}{\mbox{kg}}
\newcommand{\newton}{\mbox{N}}
\newcommand{\nanonewton}{\mbox{nN}}
\newcommand{\pascal}{\mbox{Pa}}
\newcommand{\kilopascal}{\mbox{kPa}}
\newcommand{\joule}{\mbox{J}}
\newcommand{\watt}{\mbox{W}}
\newcommand{\picowatt}{\mbox{pW}}
\newcommand{\Wmsq}{\watt/\meter^2}
\newcommand{\kilohertz}{\mbox{kHz}}
\newcommand{\megahertz}{\mbox{MHz}}

\newcommand{\Kelvin}{\mbox{K}}

\title{Acoustic Power Management by Swarms of Microscopic Robots}
\author{Tad Hogg\\
{\small Institute for Molecular Manufacturing}\\{\small Palo Alto, CA 94301}
}

\begin{document}
\maketitle

\begin{abstract}

Microscopic robots in the body could harvest energy from ultrasound to provide on-board control of autonomous behaviors such as measuring and communicating diagnostic information and precisely delivering drugs. This paper evaluates the acoustic power available to micron-size robots that collect energy using pistons. Acoustic attenuation and viscous drag on the pistons are the major limitations on the available power. Frequencies around 100kHz can deliver hundreds of picowatts to a robot in low-attenuation tissue within about 10cm of transducers on the skin, but much less in high-attenuation tissue such as a lung. However, applications of microscopic robots could involve such large numbers that the robots significantly increase attenuation, thereby reducing power for robots deep in the body. This paper describes how robots can collectively manage where and when they harvest energy to mitigate this attenuation so that a swarm of a few hundred billion robots can provide tens of picowatts to each robot, on average.

\end{abstract}

\section{Introduction}

As implanted medical devices become smaller, more numerous and more capable, they will enable many high-precision applications~\cite{dong07,nelson10,sahoo07,schulz09}. 
In particular, microscopic robots have the potential to provide precise treatments throughout the body on the scale of individual cells.
A significant challenge for realizing this possibility is providing power to the robots~\cite{amar15,bazaka13}. 
Among the options for powering such robots~\cite{cook-chennault08,freitas99,martel07,somasundar21}, ultrasound has some appealing advantages. These include noninvasive power from transducers on the skin without requiring tethers to the robots, available ultrasonic transducers in both clinical settings and for extended use~\cite{wang22}, and the ability to direct power to specific locations in the body. In addition to delivering energy, ultrasound can aid robots' tasks by altering their local environments, such as selectively opening the blood-brain barrier for drug delivery~\cite{konofagou17,pascal20}. 

One approach to acoustic power creates structured fields that apply forces to objects in the body to move them in specific directions~\cite{ghanem20}. This capability extends to microscopic robots~\cite{mohanty20,rao15,shaglwf19}, such as micron-size particles in the bloodstream~\cite{avila18}. 
For example, sound can generate oscillations in bubbles embedded in robots to propel them~\cite{aghakhani20,ren19,luo21}.
In these cases, ultrasound provides both power and control of robot behavior. 
These uses of ultrasound to exert forces on objects in the body contrast with using ultrasound to create images based on variation in tissue acoustic properties.

More complex microscopic robots could behave autonomously by using on-board logic to respond to their local environments~\cite{brooks20}. These robots could include pumps to collect or release chemicals, motors to move flagella for locomotion or vibrate the robot's surface for acoustic communication, or springs for energy storage. 

Using their internal controllers, autonomous robots determine when and how to use harvested energy, e.g., for  locomotion, communication or drug release. These decisions can account for sensed data, information received from neighboring robots, and the history of such information as recorded in the robot's memory. Even relatively simple on-board control can provide a wide range of customized behaviors~\cite{webb20}. 
For example, nearby robots could release drugs simultaneously at a microscopic target location they identify with their sensors~\cite{wiesel-kapah16}. This would produce a large sudden increase in drug concentration at that location.

In addition to power, ultrasound can provide high-resolution functional imaging~\cite{norman21} and commands to the robots by modulating the acoustic wave. This imaging could monitor the robots on scales considerably larger than an individual robot and guide robots to macroscopic target areas within the body. For example, acoustic power and activation commands could be directed toward those target areas once imaging determines a sufficient number of robots have arrived in the area. A robot could combine these external signals with those communicated from its neighbors to provide larger-scale context to that robot's sensor measurements. This combination of information from multiple scales shared among the robots could increase the precision and coordination of the robots.

Of particular interest for medicine is the eventual development of autonomous robots a few microns in size. These are the largest robots that can travel throughout the circulatory system, whose capillaries are several microns in diameter~\cite{freitas99}. 
By accessing the full circulatory system, such robots could pass as close as blood does to cells throughout the body. 
Specifically, to provide nutrients to cells and remove their waste products, capillaries carry blood within tens of microns of cells, which is close enough to allow rapid chemical exchange between the blood and the cells via diffusion~\cite{feher17}. 
Thus injecting the robots into the blood allows them to reach locations close to cells in microenvironments whose sensed properties match the robot task requirements.

Micron-size robots are considerably smaller than sub-millimeter micromachines~\cite{brooks20} based on microelectromechanical systems (MEMS) which are too large to travel through capillaries. On the other hand, they are much larger than nanoparticles used for drug delivery, which can only incorporate a few logic operations to autonomously determine their behaviors~\cite{douglas12}. 
Theoretical studies suggest micron-size robots are large enough to have sufficient computation, sensing and communication to coordinate complex activities with single-cell resolution~\cite{freitas99,hogg06a,li17,morris01,somasundar21}. Thus this size range could provide a useful combination of robot capability and access to cells via the body's circulatory system.

Microscopic robot applications can involve large numbers of robots. For instance, micron-size robots could provide customized medical diagnostics and treatments to individual cells throughout the body~\cite{jager00}. This task requires a correspondingly large number of robots to provide a reasonable treatment time, e.g., a few hours. Thus the scenarios discussed in this paper consist of billions of robots, with a total mass in the range of tens to hundreds of milligrams~\cite{freitas99}.

Interactions among robots can create swarms with performance and robustness beyond that of individual robots~\cite{floreano21}, including a variety of coherent structures and behaviors~\cite{vicsek95}. For microscopic robots, external fields can induce these interactions from forces on the robots~\cite{mohanty20} to form, for instance, specific shapes~\cite{xie19} and fluid motions~\cite{schuerle19,hernandez05}.
For more flexibility, interactions among autonomous robots can arise from their controllers coordinating with neighbors by exchanging information~\cite{rubenstein14,bonabeau99}. 

With respect to acoustic power, a swarm can consist of so many robots that they significantly alter sound propagation in tissue. In particular, robots absorbing acoustic energy increase attenuation of the sound passing through tissue containing the robots. This attenuation can be much larger than that due to the tissue, even if the robots, in total, occupy a relatively small fraction of the tissue volume. In this case, the swarm becomes the main determinant of acoustic energy propagation. Thus instead of a fixed attenuation determined by the tissue, robots that dominate the attenuation have an opportunity to improve task performance by coordinating their behaviors so as to adjust the amount and location of acoustic power.

To quantify how large numbers of microscopic robots affect acoustic power, this paper considers micron-size robots that absorb energy to power internal mechanisms. 
Ultrasound power often uses piezoelectric energy harvesters~\cite{wang07,shi16}.
Piezoelectric materials convert pressure variations to electricity. The triboelectric effect can also generate electric power for implanted devices from ultrasound~\cite{hinchet19}.
An alternative, and the focus of this paper, is mechanical energy harvesting~\cite{denisov11}. Mechanical harvesters avoid the need to distribute electricity within the robot, possibly with large resistive losses, and mechanisms to convert electricity to mechanical motion needed to actuate machines on the robot surface, e.g., for locomotion~\cite{martel08,hogg14}, or within the robot, e.g., mechanical computers~\cite{merkle18,kim20}.
The effectiveness of mechanical energy harvesters is limited by their material properties, such as friction drag.  
To estimate the ultimate potential of ultrasound power for microscopic robots, this paper considers harvesters built from atomically precise materials, whose properties and manufacture have been theoretically studied in the context of microscopic robots~\cite{alemansour20,dong07}. 
Specifically, this paper considers spherical robots that harvest energy with pistons on their surfaces, as illustrated in \fig{pistons on robot}. In addition to pistons shown in the figure, the robot contains machines that utilize the collected energy, including control and communication to allow coordinating with other robots. Aside from the pistons, the robots are taken to be stiff structures with negligible response to the acoustic pressures considered here.

\begin{figure}
\centering
\begin{tabular}{cc}
\includegraphics[width=\smallfigwidth]{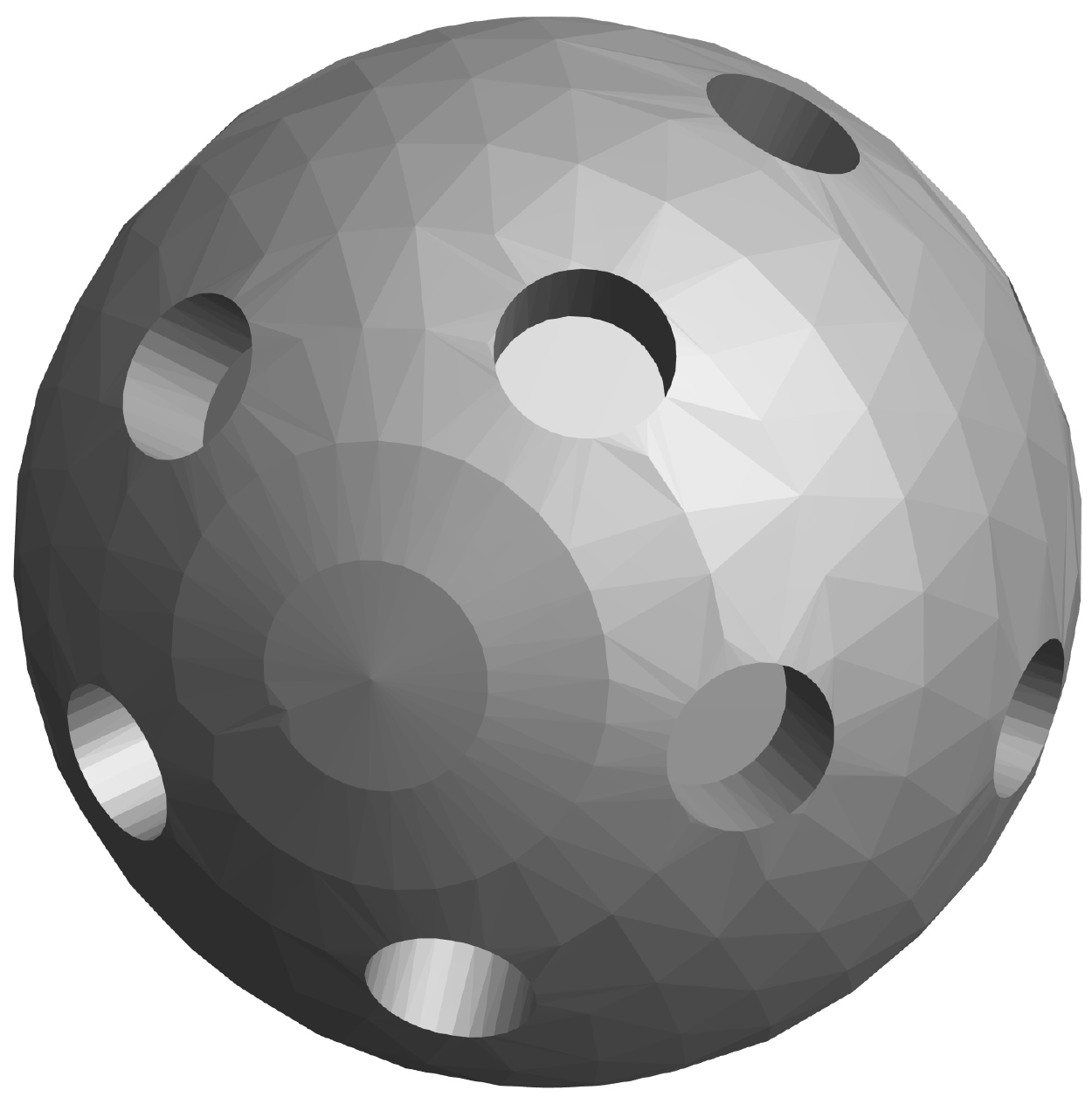}	& \includegraphics[width=\smallfigwidth]{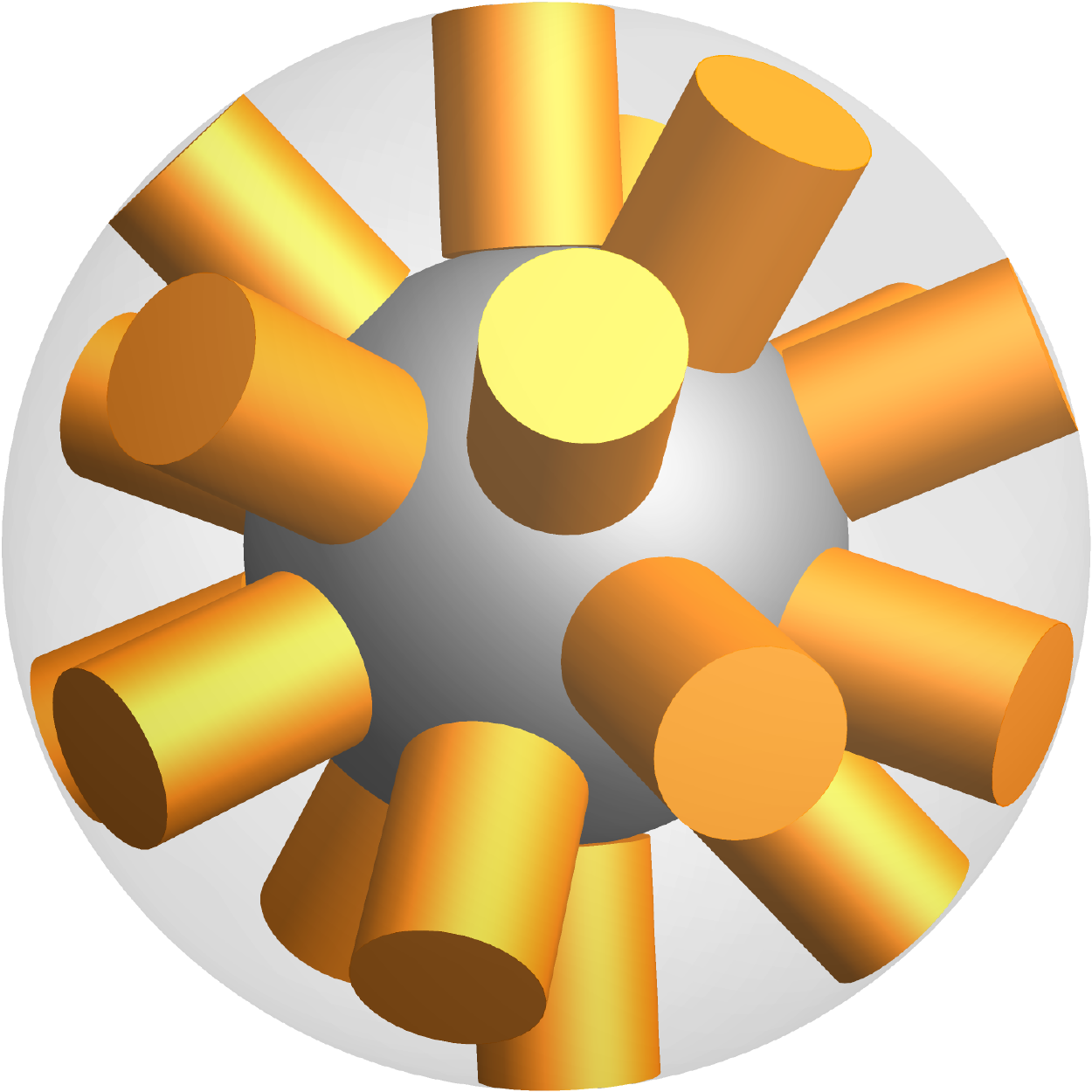} \\
\end{tabular}
\caption{Schematic illustration of the microscopic robots considered in this paper, which use pistons to collect acoustic energy. Left: robot with fully retracted pistons. Right: robot with transparent surface to show the pistons and their housings within the robot. The inner sphere indicates the portion of the robot volume under the pistons and their housings. The example used to illustrate the power available to the robots has a diameter of two microns with piston dimensions given in \tbl{piston parameters}. }\figlabel{pistons on robot}
\end{figure}

The remainder of this paper first describes sound propagation in tissue, which determines how much sound energy reaches a robot inside the body. The following section evaluates how much power a single robot can extract from this sound using pistons.
The paper then determines the increased attenuation due to a swarm of robots extracting energy from sound. Thus, in contrast to evaluating power for only one or a few microscopic robots, this paper quantifies how large numbers of robots affect the available power, and how these swarms can collectively mitigate those effects. 
Based on this analysis, the paper concludes with a discussion of the feasibility of acoustic power for swarms of microscopic robots.

\section{Acoustics in Tissue}\sectlabel{acoustics}

Sound consists of longitudinal waves propagating through materials~\cite{rienstra10,tole05,ziskin93}. 
This paper considers robots in tissue with representative acoustic properties given in \tbl{tissue parameters} and the range of frequencies shown in \tbl{sound waves}.
Even the highest frequencies have wavelengths much larger than the robot size.
For these frequencies, acoustic attenuation in tissue is proportional to frequency and is conveniently quantified  by the amplitude absorption coefficient  \absorptionTissue, given in \tbl{tissue absorption coefficients} for representative low- and high-absorption tissues~\cite{freitas99}. 
Acoustic pressure of frequency $f$ decreases by a factor of $\exp(-\absorptionTissue f x)$ over a distance $x$. Acoustic power, which is proportional to the pressure squared, decreases twice as rapidly.

Since tissue attenuation increases with frequency, powering devices deep in the body requires frequencies below about $1\,\megahertz$. On the other hand, the frequencies should be high enough to not be audible. These considerations lead to the range frequencies given in \tbl{sound waves}.

\begin{table}[tp]
\begin{center}
\begin{tabular}{lcl}
speed of sound	&\soundSpeed	&$1500\,\meter/\second$ \\
density		& \density		&$10^3 \,\kilogram/\meter^3$	\\ 
\hline \multicolumn{3}{c}{amplitude absorption coefficient}\\
soft tissue		& \absorptionTissue &$8.3/\megahertz/\meter$ \\
lung			& \absorptionTissue & $470/\megahertz/\meter$
\end{tabular}
\end{center}
\caption{Representative acoustic properties of tissue.}\tbllabel{tissue parameters}
\tbllabel{tissue absorption coefficients} 
\end{table}

\begin{table}[tp]
\begin{center}
\begin{tabular}{ccc}
frequency	& wavelength 	& $k \; \rRobot$\\ \hline
$20\,\kilohertz$		& $75\,\millimeter$	& $8\times 10^{-5}$\\
$100\,\kilohertz$	& $15\,\millimeter$	& $4\times 10^{-4}$\\
$1000\,\kilohertz$	& $1.5\,\millimeter$	& $4\times 10^{-3}$\\
\end{tabular}
\end{center}
\caption{Sound frequencies and wavelengths compared to the robot size: $k=2\pi/\lambda$ is the wave number of sound with wavelength $\lambda=c/f$ at frequency $f$, and $\rRobot = 1\,\micron$ is the robot's radius. 
}\tbllabel{sound waves}
\end{table}

Ultrasound imaging typically uses a single transducer covering a small portion of the skin. For powering devices throughout the body, multiple transducers distributed over the body ensure that each device is relatively close to a transducer. 
Specifically, this study evaluates power available within about $20\,\centimeter$ of the nearest transducer. This corresponds to using multiple transducers to cover the body, or at least the part of the body where robots require power to perform their tasks.

\section{Extracting Energy from Acoustic Pressure Waves}

\begin{table}[tp]
\begin{center}
\begin{tabular}{lcl}
\multicolumn{3}{c}{piston}\\
piston diameter			&$d$			&$300\,\nanometer$ \\
piston thickness		&$\tau$		&$10 \,\nanometer$	\\
piston range of motion	&$2 a$		&$200 \, \nanometer$	\\
piston cross section area	&$A=\pi (d/2)^2$	&$0.071 \, \micron^2$	\\
piston sliding area		&$\ASlidingPiston=\pi d\tau$	&$0.019 \, \micron^2$	\\
\hline
\multicolumn{3}{c}{constant-force spring}\\
perpendicular overlap	&$\Lspring$			&$\approx 35\pm10\,\nanometer$ \\
parallel overlap			&$\hspring$			& up to $\approx 200\,\nanometer$ \\
\hline
\multicolumn{3}{c}{housing}\\
housing diameter		&$D$			&$340\,\nanometer$ \\
housing depth			&$H$			&$440\,\nanometer$ \\
\end{tabular}
\end{center}
\caption{Geometry of a piston, constant-force spring and housing. Overlaps for the constant-force spring vary with piston position and ambient pressure (see \fig{piston constant force}).}\tbllabel{piston parameters}
\end{table}

\begin{figure}
\centering
\includegraphics[width=\widefigwidth]{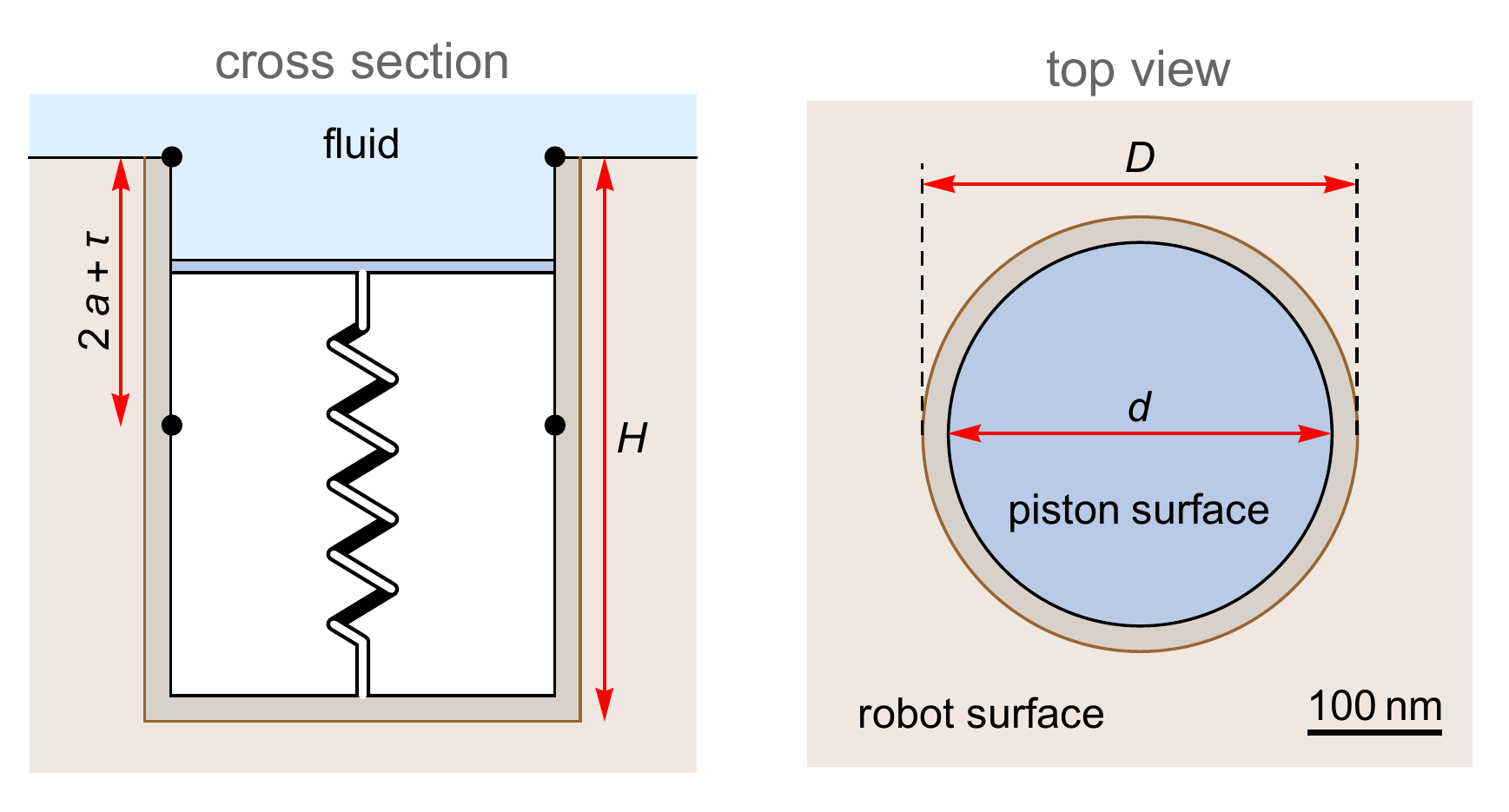}
\caption{Cross section and top views of a piston (gray) and its housing (dark brown) in a robot. In the cross section, the four black points indicate the limits of piston motion, i.e., the upper limit for the top of the piston and the lower limit for the bottom of the piston. The distance between these limits is the sum of the range of motion, $2 a$, and the piston thickness, $\tau$. The spring under the piston provides a restoring force. In the top view, the scale bar corresponds to values in \tbl{piston parameters}.
}\figlabel{piston geometry}
\end{figure}

Pistons moving in response to changing pressure can extract energy from acoustic waves.
\fig{piston geometry} shows the geometry of a piston, constrained to move over a limited range within a housing. The figure does not include linkages from the piston to mechanisms within the robot powered by the piston's motion.

Sound consists of pressure variations around an ambient value. The ambient pressure varies with time and position in the body~\cite{freitas99}. For example, a robot moving with the blood encounters a decrease in pressure of about $20\,\kilopascal$ over about $10\,\second$ as it moves from arteries to veins. In addition, pressure in arteries varies by about $5\,\kilopascal$ during each heartbeat.

A piston with a fixed restoring force would respond to acoustic pressure only when the force from the ambient pressure is close to that restoring force. Outside that range, the piston would be pinned at one of its limits of motion and not collect energy. To avoid this limitation, the restoring force must adjust to changing ambient pressure. 
Since ambient pressure changes much more slowly than acoustic variation, the force could be adjusted based on the average pressure on the piston measured over multiple periods of the acoustic wave. This measurement could use force sensors on the piston mechanism or pressure sensors elsewhere on the robot surface~\cite{freitas99}. Alternatively, the controller could measure the average position of the piston and adjust the restoring force to keep the average near the middle of the piston's range. For example, a controller could adjust the restoring force by averaging over several milliseconds to adapt to changes in ambient pressure.

The remainder of this section describes a mechanism to adjust to ambient pressure changes, the resulting piston motion, and the power extracted from acoustic pressure waves.

\subsection{Constant-Force Springs}\sectlabel{constant force springs}

One approach to compensating for ambient pressure is a conventional spring, in which restoring force is proportional to displacement. However, the increasing force with displacement stores much of the energy from an increasing acoustic pressure in the potential energy of the spring. Without additional mechanisms to capture that energy, the potential energy would be returned to the fluid when the pressure decreases, leading to scattering rather than absorption of acoustic energy.

\begin{figure}
\centering
\includegraphics[width=\mfigwidth]{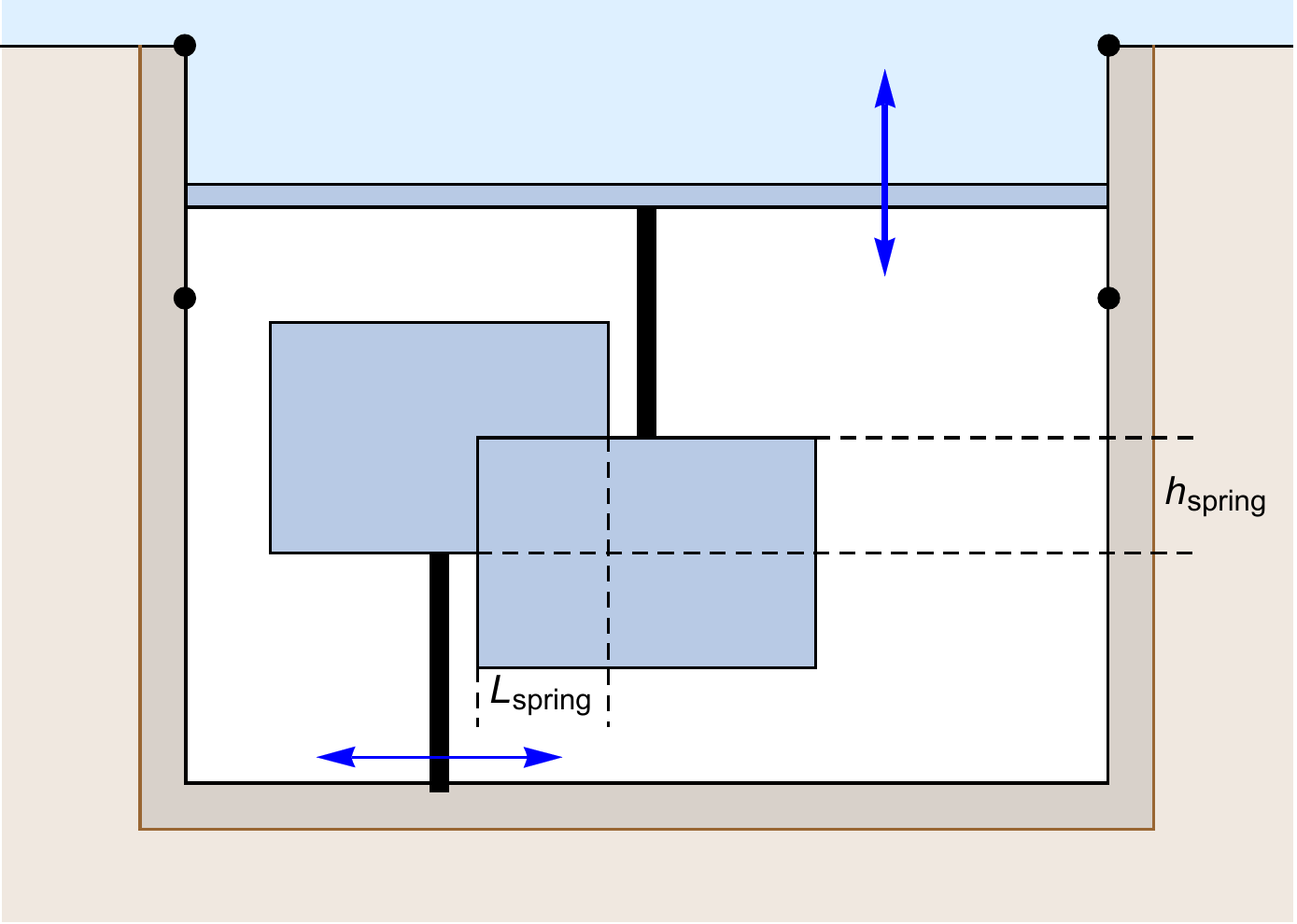}
\caption{Overlapping sheets forming an adjustable constant-force spring. The thick vertical lines connect the sheets to the piston or housing. These connections could be offset from the plane of overlap between the sheets to avoid limiting the sheets' motions. The lower sheet moves vertically with the piston (indicated by the vertical arrow), thereby changing the vertical overlap, \hspring, between the sheets.  Moving the position of the upper sheet's attachment to the housing (indicated by the horizontal arrow) adjusts the horizontal overlap, \Lspring. Horizontal scale is exaggerated compared to \fig{piston geometry}.
}\figlabel{piston constant force}
\end{figure}

Constant-force springs are a better way to adjust for ambient pressure changes. For example, a pre-compressed conventional spring with applied forces small compared to that compression has relatively small displacements around the compressed location. In this case, Hooke's Law gives nearly constant force over the range of displacements.

At the nano scale, overlapping molecular devices can behave as constant-force springs due to van~der~Waals interactions~\cite{cumings00,liu12}. 
The force is in the direction of increasing overlap. This leads to the geometry in \fig{piston constant force}, where the sheet attached to the piston is below the horizontally-adjustable sheet attached to the housing. In this configuration, the spring pushes upwards, against the pressure on the piston from the fluid.
The force is proportional to the length of overlap, \Lspring, between the sheets in the direction perpendicular to the motion:
\begin{equation}\eqlabel{F spring}
\Fspring = \Kspring \Lspring
\end{equation}
Unlike Hooke's law, this force is independent of the position in the direction of motion, \hspring.
The spring constant \Kspring\ is $0.16\,\newton/\meter$ for nested nanotubes~\cite{cumings00} and $0.2\,\newton/\meter$ for graphene sheets~\cite{liu12}. Sheets are convenient for this application because a horizontal change in their position, i.e., changing \Lspring, allows adjusting the spring's force in response to changing ambient pressure.

For example, the force on a piston with area given in \tbl{piston parameters} due to one atmosphere of external pressure is $7\,\nanonewton$. The constant-force spring compensates for this force with an overlap $\Lspring = 35\,\nanometer$. This overlap is considerably less than the piston diameter, allowing more than enough room for the overlapping sheets within the piston housing.

In the geometry of \fig{piston constant force}, the vertical extent of each sheet must be at least equal to the range of piston motion. This limits the piston's range to less than a third of the housing depth. More efficient use of housing volume is possible because horizontal overlaps several times larger than $35\,\nanometer$ can fit within the housing diameter. Thus the constant-force spring could connect to the piston via a lever: a larger overlap would increase the spring's force enough to compensate for pressure on the piston while the vertical overlap of the sheets changes by a corresponding multiple smaller than the piston's movement. Thus the sheets could have smaller vertical extent, and the piston could move through a greater fraction of the housing, than in \fig{piston constant force}. This would allow a greater range of piston motion in a given housing depth, or reducing the housing depth required for a given range of motion, thereby reducing the housing volume and providing more space in the robot for other devices.

The constant-force spring in \fig{piston constant force} consists of two sheets sliding over each other. These sheets could be much thinner than the piston diameter, so the housing could contain multiple sliding sheets with separate actuators to adjust their overlaps independently. This would provide redundancy against failures, or allow different motors to handle adjustments over different ranges of forces. Alternatively, the force could come from combining one or more adjustable planes with nested nanotubes, whose forces are not adjustable. In this case, the nanotubes could provide the force for the lowest ambient pressure in the body, while the adjustable planes provide forces over the range of ambient pressure above this minimum value.

\subsection{Piston Friction}\sectlabel{piston drag}

At the small sizes considered here, frictional drag is the dominant force on robot mechanisms. Thus, assessing the feasibility of mechanical energy harvesting must account for friction, which depends on surface size and structure~\cite{hsu14}. Specifically, drag force on a piston moving at speed $v$ is
\begin{equation}\eqlabel{F drag}
\Fdrag = \kTotal v
\end{equation}
with the drag coefficient 
\begin{equation}\eqlabel{k drag}
\kTotal =  \kFriction + \kLoad
\end{equation}
having contributions from internal friction of piston motion, \kFriction, and from powering mechanisms within the robot, \kLoad.
Contributions to \kLoad\ include, for instance, friction within robot components such as mechanical computers~\cite{merkle18}, driven by a piston's motion.

Internal friction consists of viscous drag from motion through the fluid around the robot, \kViscous, and sliding friction between the piston and its housing, which, for atomically smooth surfaces, has the form $\kSliding \ASliding$~\cite{drexler92}, where \kSliding\ is the sliding drag coefficient and \ASliding\ is the area of the sliding surfaces. Thus
\begin{equation}\eqlabel{k drag internal}
\kFriction 	= \kViscous + \kSliding \ASliding
\end{equation}
Friction arising from motion of linkages connecting the piston to devices within the robot are considered part of dissipation external to the piston included in \kLoad.
The remainder of this section estimates the magnitude of these contributions to \kFriction.

\subsubsection{Viscous Drag}\sectlabel{piston viscous drag}

\begin{figure}
\centering
\includegraphics[width=\widefigwidth]{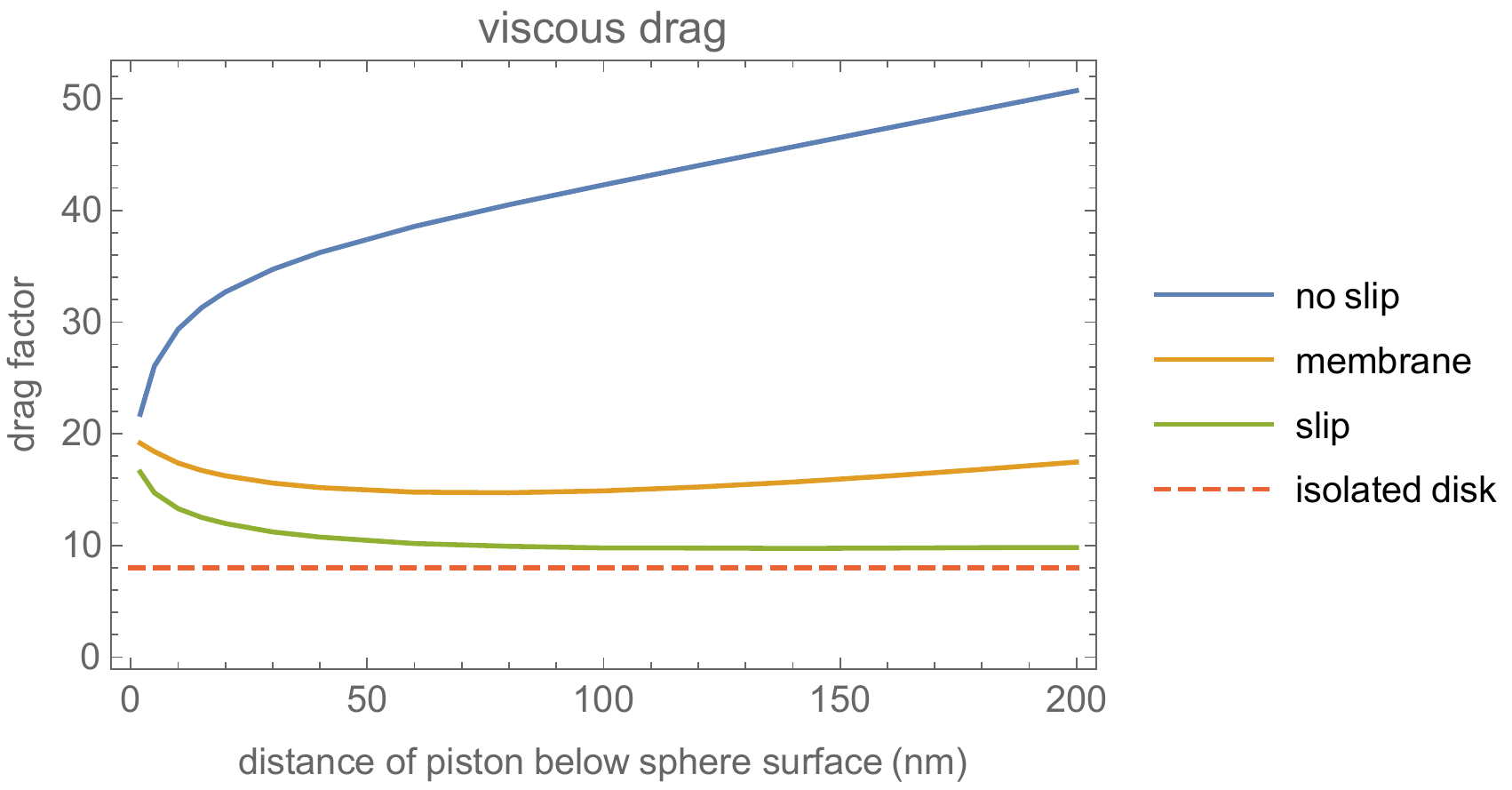}
\caption{Viscous drag factor, $g = \kViscous/(d \viscosity)$, for the piston described in \tbl{piston parameters} as a function of its position below the robot surface for the boundary conditions described in the text. The dashed line is the value for an isolated disk moving face-on through the fluid. 
}\figlabel{piston drag}
\end{figure}

The first contribution to \kFriction\ in \eq{k drag internal} arises from viscous drag in the fluid.
An object moving at speed $v$ through a fluid with viscosity $\viscosity$ at low Reynolds number experiences a drag force proportional to its size and the fluid viscosity~\cite{happel83}. A convenient characterization of this drag is the viscous drag factor $g = \kViscous/(d \viscosity)$, where $d$ is the object's size.The value of $g$ depends on the object's shape and orientation, as well as any nearby boundaries. 
For instance, in unbounded fluid, a sphere with diameter $d$ has $\Fdrag = 3 \pi \viscosity d v$~\cite{berg93}, so $g=3\pi$. 
Similarly, a flat disk with diameter $d$ moving face-on through the fluid has $g = 8$~\cite{berg93}. 

Numerically evaluating the fluid flow gives the viscous drag as a function of the piston's position within its housing.
The drag depends on how the fluid interacts with the housing surface; in particular, the deviation from the no-slip boundary condition between fluids and solid surfaces~\cite{mate08,squires05}.
The fluid motion considered here has large velocity gradients near the edge of the moving piston, which can lead to some slip~\cite{granick03,lauga07}. While direct experimental measurements of the slip are difficult, molecular dynamics simulations identify factors affecting slip~\cite{arya03}, though such computations are generally restricted to short distances and time scales and use shear rates considerably larger than the flows considered here. Factors affecting the boundary condition include the structure of the fluid on nanometer length scales and its composition, particularly the amount of dissolved gases. A precise estimate of the boundary conditions and hence the piston drag will require experimental investigation. 

Due to this range of possibilities, this study evaluates the drag with several boundary conditions on the fluid's motion at the housing wall: 1) no-slip except for slip within $1\,\nanometer$ of the piston's surface due to the high shear between the stationary housing and moving piston, 2) an elastic membrane between housing and piston so that fluid speed changes linearly between them, and 3) slip along the housing wall, e.g., due to suitably engineered surface or an approximation to the reduced slip at high shear from dissolved gases. 
\fig{piston drag} shows the drag factor $g$ for a piston with parameters given in \tbl{piston parameters} with these choices. For comparison, the dashed line shows the drag on an isolated disk moving face-on through the fluid far from any boundaries.

Based on \fig{piston drag}, this study uses a relatively large value, $g=45$, and, for simplicity, takes the value to be independent of piston position. This gives $\kViscous = g d \viscosity = 1.4\times 10^{-8}\,\kilogram/\second$ for the value of \viscosity\ given in \tbl{fluid parameters}.
In practice, the housing surface may allow more slip at the boundary, so this choice approximates a lower bound on acoustic power.

\subsubsection{Sliding Drag}\sectlabel{piston sliding drag}

The second contribution to \kFriction\ in \eq{k drag internal} is friction from sliding surfaces. 
For stiff materials, theoretical estimates for atomically-flat surfaces give \kSliding\ as somewhat less than $10^3\,\kilogram/(\meter^2 \second)$~\cite{drexler92,freitas99} for speeds well below the speed of sound, as is the case for piston motion. 
Thus $\kSliding = 10^3\,\kilogram/(\meter^2 \second)$ is used here to estimate an upper bound for sliding friction.

The sliding area includes the surface at the edge of the piston next to the housing and the overlapping surface of the constant-force spring:
\begin{equation}
\ASliding = \ASlidingPiston + \Lspring \hspring
\end{equation}
The overlap area of the constant-force spring, $\Lspring \hspring$ is less than $0.01\,\micron^2$, so \tbl{piston parameters} gives $\ASliding < 0.03\,\micron^2$.
Thus  $\kSliding \ASliding < 3\times 10^{-11}\,\kilogram/\second$ is much less than the viscous drag estimate in \sect{piston viscous drag}, i.e., $\kFriction \approx \kViscous$.

The above discussion estimates the drag on the piston as it moves in response to acoustic pressure variation.
In addition, actuators alter the overlap \Lspring\ of the constant-force spring in response to changes in ambient pressure.
The sliding drag during this adjustment is
\begin{equation}\eqlabel{F drag for spring}
\FdragSpring = \kSliding (\Lspring \hspring)  \frac{d\Lspring}{dt}
\end{equation}
An example is the changing ambient pressure during a heartbeat of about $p=5\,\kilopascal$ in one second. This requires changing \Lspring\ by $2\,\nanometer$.
Using the bound on sliding area described above, i.e, $\Lspring \hspring < 0.01\,\micron^2$, the dissipation due to friction during this change is negligibly small, less than $10^{-16}\,\picowatt$. The small size of this value is due to the relatively long time over which ambient pressure changes, and the consequent slow speed required to adjust the overlap of the constant-force spring.

Actuators adjusting the horizontal position of the surfaces apply forces comparable to $F = \Kspring \hspring$ to change the overlap \Lspring. To reduce overlap by $\Delta \Lspring$, the actuator does work $W = F \Delta \Lspring$. The overlap \hspring\ varies with the piston's position. A typical example for the geometry of \fig{piston constant force} is $\hspring = 100\,\nanometer$ and $F=20\,\nanonewton$. Theoretical estimates indicate that, for example, electrostatic actuators~\cite{drexler92} could provide such forces and be small enough to adjust overlaps for the piston sizes considered here.
Continuing with the previous example of $5\,\kilopascal$ pressure change in one second, $\Delta\Lspring = 2\,\nanometer$ so $W =4\times 10^{-17}\,\joule$.
The work done by the actuator is stored as potential energy in the position of the sheets. This energy could be recovered as work done by the sheets to increase \Lspring\ when ambient pressure later increases. However, even if this work were entirely dissipated during the one second change in ambient pressure, it amounts to a still negligible dissipation of $10^{-5}\,\picowatt$. This potentially avoidable dissipation is much larger than that due to sliding friction described in \eq{F drag for spring}, a property also seen in rotary motion of atomically precise mechanisms~\cite{hogg17}.

\subsection{Piston Motion}\sectlabel{piston motion}

For the position of the piston, $x$, define $x=0$ to be the middle of the piston's range, and pick the sign of $x$ so positive values mean the piston is farther from the center of the robot. Let $a>0$ be the maximum value of $x$, so the piston ranges between $-a$ and $a$.

The piston moves in response to forces from the applied pressure, the spring, \Fspring, and damping, \Fdrag. The pressure is $\Pambient + p(t)$ where $p(t)$ is the acoustic pressure, and \Pambient\ is the ambient fluid pressure.  
Fluid pressure pushes inward and the spring pushes outward. The piston stops whenever these forces attempt to push it beyond its range of motion. Thus, unless the piston is stopped at a limit to its range of motion, its position $x$ changes according to
\begin{equation}\eqlabel{piston motion with inertia}
m \frac{d^2 x}{d t^2} = -\Fdrag + \Fspring - \left( \Pambient + p(t) \right) A
\end{equation}
where $A$ is the cross section area of the piston and $m$ its mass. 

Viscous forces dominate the motion of objects of the size and speeds considered here~\cite{purcell77}.
This means that the piston moves at the speed at which \Fdrag\ balances the forces applied to the piston. That is, the piston moves at its terminal velocity in the fluid, so \eq{piston motion with inertia} becomes
\begin{equation}\eqlabel{piston motion}
\kTotal \frac{d x}{d t} = \Fspring - \left( \Pambient + p(t) \right) A
\end{equation}
by using \eq{F drag}.
If the piston reaches a limit to its motion, at $x=\pm a$, it remains there, i.e., $dx/dt=0$, until the applied force changes sign.

The constant-force spring adjusts to the ambient pressure, so that $\Fspring = \Pambient A$.
We arbitrarily pick the origin of time at the minimum acoustic pressure, so $p(t) = -p \cos(\omega t)$ for sound with angular frequency $\omega = 2\pi f$ and amplitude $p > 0$ at the robot's location.
Convenient dimensionless parameters for describing the piston motion are
\begin{equation}\eqlabel{motion parameters}
\begin{split}
\lambda 		&\equiv \frac{p A}{a \omega \kFriction } \\
\kLoadRatio 	&\equiv \frac{\kLoad}{\kFriction}
\end{split}
\end{equation}
$\lambda$ is a ratio of acoustic pressure and internal drag forces on the piston.
Defining normalized position $X=x/a$ and time $\tau=\omega t$, \eq{piston motion} becomes 
\begin{equation}\eqlabel{normalized piston motion}
\frac{d X}{d \tau} = \frac{\lambda}{1+\kLoadRatio} \cos(\tau)
\end{equation}

Viscous drag on the piston depends on its position (see \fig{piston drag}). For simplicity, we ignore this variation and instead consider the drag coefficient \kViscous, and hence \kFriction, to be independent of position, as described in \sect{piston viscous drag}. 
In this case, \eq{normalized piston motion} gives
\begin{equation}\eqlabel{piston displacement}
X(\tau)= \frac{\lambda}{1+\kLoadRatio} \sin(\tau)
\end{equation}
when the piston starts at the center of its range of motion, i.e., $X(0)=0$.

\begin{figure}
\centering
\includegraphics[width=\widefigwidth]{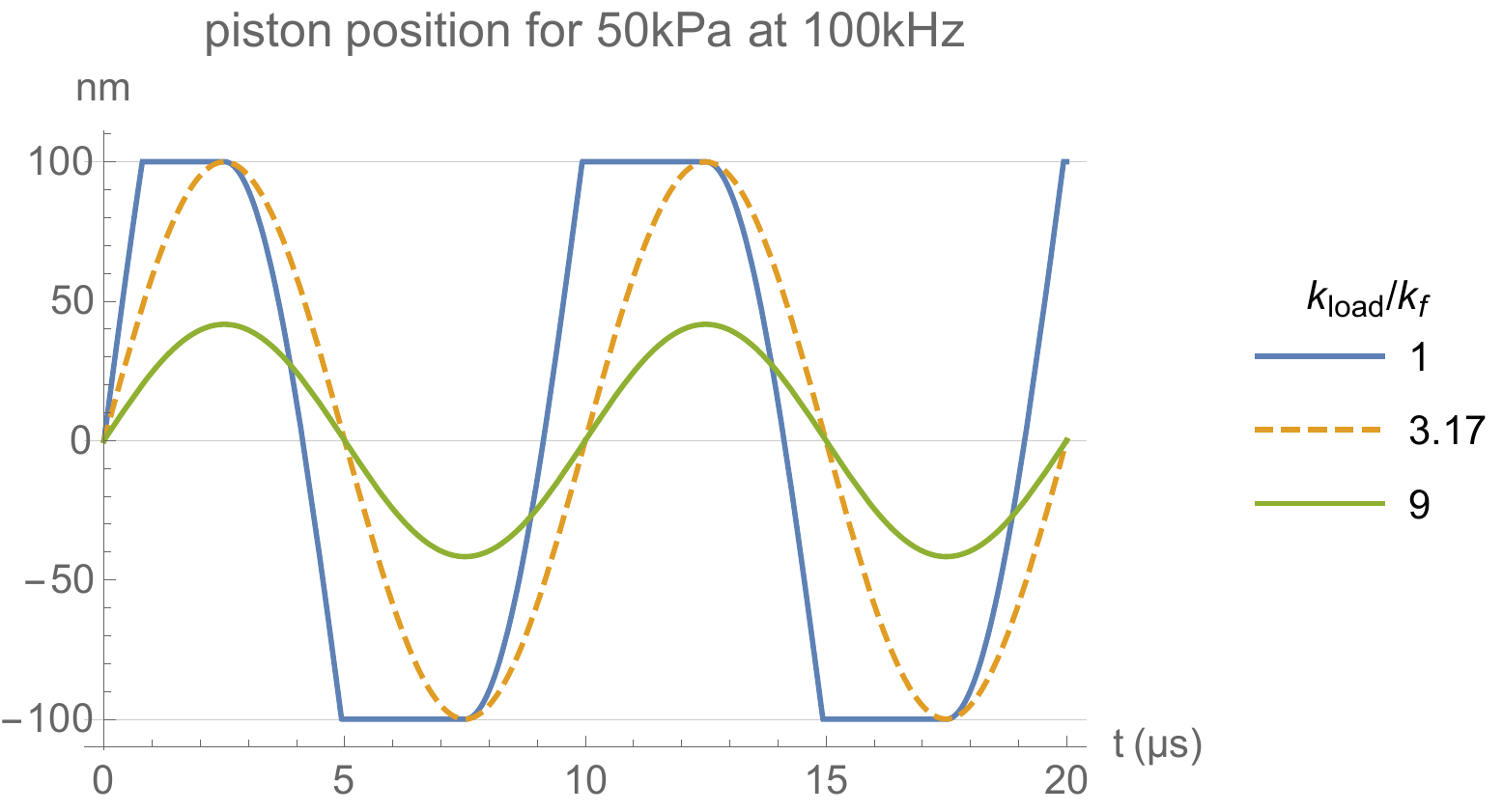}
\caption{Motion of pistons described in \tbl{piston parameters} in response to acoustic pressure of $50\,\kilopascal$ at $100\,\kilohertz$. In this case, \eq{motion parameters} gives $\lambda = 4.17$. The curves show the piston's position as a function of time with the piston connected to one of three loads: \kLoad\ equal to $\kFriction$, $(\lambda-1)\kFriction$ (dashed) or $9\kFriction$. The dashed curve is the smallest load for which \eq{piston displacement} describes the complete motion.
}\figlabel{piston motion}
\end{figure}

\eq{piston displacement} describes the complete piston motion when $\lambda \le 1+\kLoadRatio$. 
Otherwise, the piston spends part of each cycle stopped at the limits of its range, and \eq{normalized piston motion} only holds between these stops.
\fig{piston motion} illustrates these cases.

\subsection{Piston Power}\sectlabel{piston power}

As the piston moves, it dissipates energy against the drag force at the rate $\Fdrag \dot{x}= \kTotal \dot{x}^2$, from \eq{F drag}. Of this amount, $\kLoad \dot{x}^2$ is dissipated by the load as power available to the robot. 

When \eq{piston displacement} describes the complete motion of the piston, the time-average power over a cycle of the acoustic pressure variation is
\begin{equation}\eqlabel{power}
\begin{split}
\PTotal 	&= \frac{1}{2} A^2 p^2 \frac{1}{\kTotal} \\
\PLoad 	&= \PTotal \frac{\kLoad}{\kTotal}	
\end{split}
\end{equation}
From \eq{k drag}, the maximum \PLoad\ occurs when $\kLoad = \kFriction$ (i.e., $\kLoadRatio=1$) in which case $\PLoad = \PTotal/2 = A^2 p^2/(8 \kFriction)$. This holds provided \eq{piston displacement} describes the complete motion, i.e., when $\lambda \le 2$.

When \kLoad=\kFriction\ and $\lambda > 2$, the piston stops at its limit of motion for a portion of each acoustic period. The piston delivers no power while stopped. In this case, the piston delivers more power to the load for somewhat larger values of \kLoad. In particular, $\kLoad=(\lambda-1)\kFriction$ is the smallest load for which \eq{piston displacement} describes the complete motion. From \eq{power}, this choice gives larger \PLoad\ than any larger value of \kLoad. 
Numerically evaluating the motion for \kLoadRatio\ between one and $\lambda-1$ shows the maximum \PLoad\  occurs at \kLoad\ slightly smaller than $\lambda-1$. I.e., the larger power from faster piston speed while it moves more than compensates for the lack of power during the short times the piston stops at its limit.
However, this maximum is only slightly larger than \PLoad\ when $\kLoadRatio=\lambda-1$: less than 3\% more for $\lambda < 5$, the range relevant for this discussion. 
Due to the minor benefit of smaller \kLoad, for simplicity we use $\kLoadRatio=\lambda-1$ to estimate the available power when $\lambda > 2$. 

Combining these cases gives
\begin{equation}\eqlabel{available power}
\PLoad = \begin{cases}
\frac{1}{8 \kFriction}A^2 p^2				& \mbox{if } \lambda \le 2 \\
\frac{1}{2} a \omega (A p - a \omega \kFriction)	& \mbox{if } \lambda > 2
\end{cases}
\end{equation}
At sufficiently high frequencies or low pressures, $\lambda \le 2$ so available power is proportional to the square of the pressure and independent of frequency. 
Conversely, at very low frequencies or high pressures (more specifically, $\lambda \gg 1$), $\PLoad \approx a \omega A p/2$, which is also $(\pi/2)p V f$ where $V=2 a A$ is the full displacement volume of the piston. In this limit, power used by the load is proportional to frequency and pressure, and is much larger than the power dissipated by viscous drag.

For example, consider a piston with the geometry of \tbl{piston parameters} receiving $50\,\kilopascal$ of acoustic pressure. At $100\,\kilohertz$ and $200\,\kilohertz$, the piston moves through its full $200\, \nanometer$ range of motion and provides $84\,\picowatt$ and $115\,\picowatt$, respectively. At $300\,\kilohertz$ and $500\,\kilohertz$ the piston delivers $116\,\picowatt$, independent of the frequency, while moving through $140\, \nanometer$  and $83\, \nanometer$, respectively.
Larger distances from the skin provide less pressure due to attenuation, as indicated in  \tbl{tissue absorption coefficients}. For instance, in soft tissue extending $20\,\centimeter$ from the skin, $500\,\kilohertz$ has pressure reduced from $50\,\kilopascal$ to $22\,\kilopascal$ and the piston provides $22\,\picowatt$ while moving through $36\, \nanometer$.

\subsection{Piston Reliability}

In response to $100\,\kilohertz$ sound, a piston oscillates $10^5$ times a second. 
Thus a piston oscillates about a billion times during a mission lasting several hours.
An important question is the reliability of pistons and their associated mechanisms for this many operations.

As an example of reliability, micromechancial machines such as MEMS mirrors readily achieve over $10^{12}$ operations~\cite{douglass98}.
For nanoscale machines, theoretical analysis suggests atomically precise components have low failure rates~\cite{drexler92}. 
This could apply to the moving components inside the piston housing, including constant-force springs and couplings to internal robot mechanisms, because those mechanisms are protected from the external biological environment.
Moreover, experimental studies of sliding in nested nanotubes show no wear at the atomic scale~\cite{cumings00}.
Thus we can expect reliable performance for mechanical devices inside the housing.

Another reliability issue is fouling of surfaces exposed to fluids. Such fouling could increase drag on the pistons, reducing the available power.
Experience with implanted micro devices shows that suitable surface design can provide biocompatibility and reduce fouling~\cite{grayson04}, suggesting fouling need not be a problem for operating times well beyond a few hours.

However, if fouling is a significant problem over the duration of a mission, modifications to the devices could reduce the problem.
For instance, the piston could be covered with a watertight membrane connected to the upper edge of the housing. This would exclude external fluids from the housing, thereby avoiding fouling along the housing sides. This approach would somewhat reduce available power because some energy would be dissipated in stretching the membrane. 
An alternative approach is to keep some pistons covered and inactive. These pistons would act as a reserve, to be uncovered as needed to replace pistons that become stuck. The covered pistons would not collect energy, thereby reducing available power. In terms of power generation, covering some pistons would be equivalent to using all pistons with a reduced duty cycle.

One other reliability issue arises from the barriers limiting piston motion. These barriers could extend inward from the housing at the indicated locations in \fig{piston geometry}. The barrier above the piston would be in the fluid and may become fouled or alter the flow along the side of the housing to increase viscous drag.
To avoid these potential problems, the barriers could be entirely inside the housing, e.g., limiting the motion of the spring mechanism rather than directly acting on the piston. 
With a constant-force spring consisting of atomically-precise structures, these barriers could also be atomically precise to provide well-defined performance without wear. However, if such barriers are atomically flat, they could stick to the part of the constant-force spring reaching the barrier due to van~der~Waals forces.
Specifically, van~der~Waals force per unit area on planar surfaces separated by distance $\delta$ is $\PvdW = \Hamaker/(6\pi \delta^3)$~\cite{drexler92,wautelet01}, with the Hamaker constant, \Hamaker, in the range $10^{-20}\mbox{--}10^{-19}\,\joule$~\cite{lauga07,wautelet01}.

As an estimate of this force, suppose the barriers consist of a set of tabs extending inward from the housing, with total area of $10\,\nanometer^2$ coming into contact with part of the spring or piston to stop its motion. If these were atomically smooth surfaces, contact with atomic spacing of about $0.3\,\nanometer$ results in van~der~Waals force of about $2\,\nanonewton$, using the upper range of values for \Hamaker.
This force is comparable to the force on the piston due to variation in acoustic pressure near the transducers (see \sect{constant force springs}).
Thus van~der~Waals forces could keep the piston stuck at its limit even when the acoustic pressure changes sign and acts to push it away from the limit. 
This would be particularly problematic when acoustic pressure variation is small, i.e., for robots deep in the body or in high-attenuation tissue. However, when the pressure variation is small, the piston will not reach the limits of its motion, as discussed in \sect{piston motion}, thereby avoiding the problem.
If necessary, adding extrusions on the surfaces so only a small portion of the surfaces come into contact~\cite{drexler92,freitas04} can avoid this difficulty.

\section{Acoustic Power for a Robot}\sectlabel{available power}

\begin{table}[tp]
\begin{center}
\begin{tabular}{cc}
\multicolumn{2}{c}{pistons}	\\
duty cycle			& $50\%$\\
number			&$20$ \\
fraction of robot volume	& $19\%$\\
fraction of robot surface	& $14\%$\\
\hline
\multicolumn{2}{c}{ultrasound source}	\\
intensity			&$1000\,\watt/\meter^2$\\
reflection loss		&$10\%$\\
source pressure	&$50\,\kilopascal$\\
\end{tabular}
\end{center}
\caption{Robot power scenario for the example shown in \fig{pistons on robot}. The fractions include the piston's housing, indicating how much of the robot these devices use. Inside the robot, the bottoms of the pistons use $46\%$ of the surface area indicated by the inner, solid sphere of the figure. The pistons themselves, i.e., the moving portion of the device, use $\fPistons = 11\%$ of the robot's surface area. The sound source intensity corresponds to $55\,\kilopascal$ (see \eq{flux}), which is reduced by the reflection loss to give the source pressure in the tissue next to the transducer.}\tbllabel{robot scenario}
\end{table}

Safety limits ultrasound intensity to about $1000\,\Wmsq$ for extended use~\cite{ng02,shankar11}, 
which corresponds to pressure amplitude $p=55\,\kilopascal$ because the time-average energy flux of a plane wave is
\begin{equation}\eqlabel{flux}
\flux = \frac{1}{2} \frac{p^2}{\density \soundSpeed}
\end{equation}
where $p$ is the amplitude of sound pressure variation, \density\ is the fluid density and \soundSpeed\ the speed of sound.
For transducers well-coupled to the skin over soft tissues, reflection losses can be fairly low, e.g., $10\%$. 
These values provide the acoustic source parameters in  \tbl{robot scenario}.
Other cases have larger losses, e.g., $40\%$ for transit through the skull~\cite{freitas99}.

A robot can use multiple pistons to extract power, as illustrated in \fig{pistons on robot}. An example is the scenario in \tbl{robot scenario}.
The duty cycle given in the table could arise from variation in sound pressure at the robot's location due to adjustments to the transducer to shift  the location of minimum pressure variation in standing waves. For example, these changes could shift the sound field by half a wavelength to ensure robots are not permanently located in a pressure minimum due to interference from strong reflections, e.g., from nearby bones.
The duty cycle could also arise from a robot occasionally turning off power collection to enable (or simplify the design or control of) other uses for its surface that are affected by piston motions. Examples include using surface vibrations for communication~\cite{hogg12} or locomotion~\cite{hogg14}. Another reason to occasionally turn off the pistons arises from their motion altering the fluid flow near the surface, and thus the pattern of fluid stresses on the robot surface. Such changes could degrade the accuracy of a robot using these stresses to estimate its position and motion~\cite{hogg18}. Turning off the pistons when performing these estimates avoids this problem.
Alternatively, if there is no need for pistons to operate with a duty cycle, the results given here correspond to a robot with $10$ pistons operating continually, i.e., $100\%$ duty cycle, thereby leaving more surface area for other uses, e.g., chemical sensing.

\begin{figure}
\centering
\begin{tabular}{cc}
\includegraphics[width=\mfigwidth]{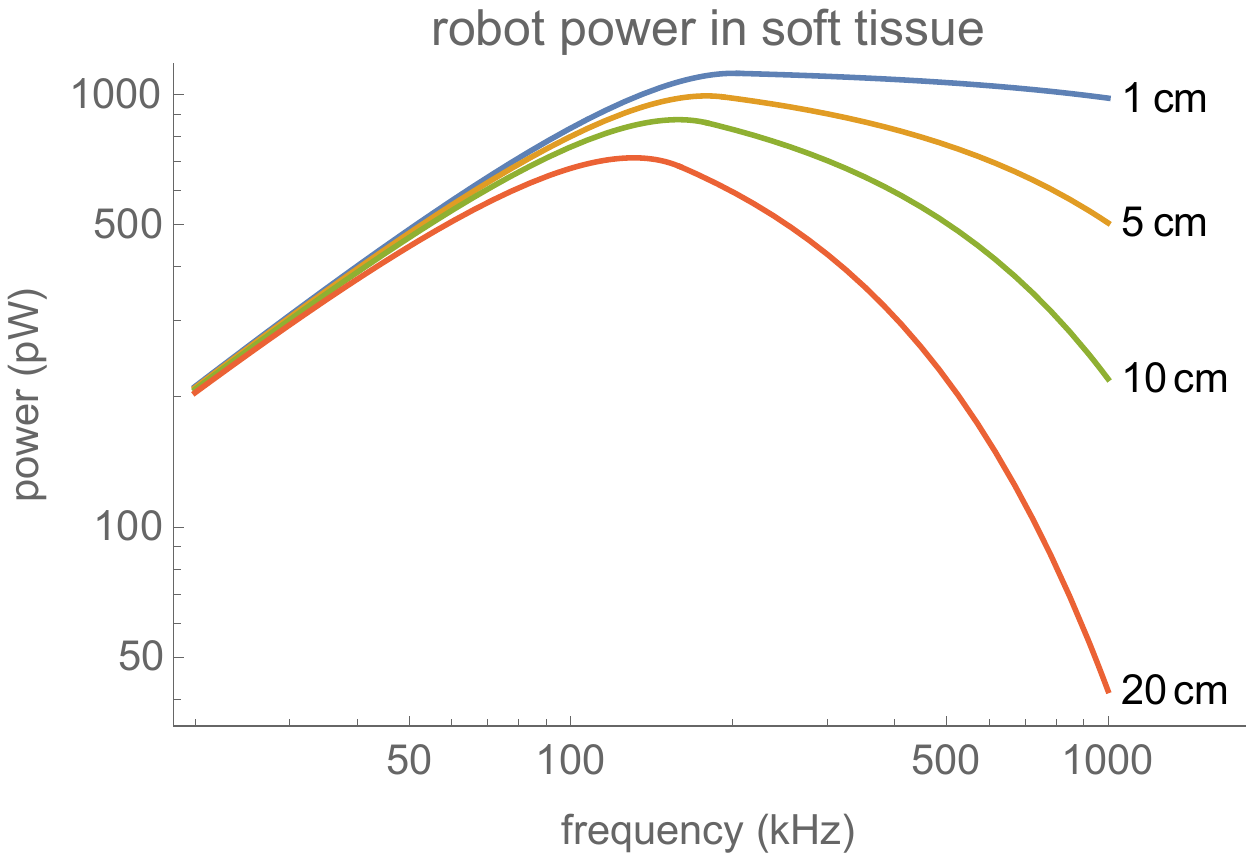} & \includegraphics[width=\mfigwidth]{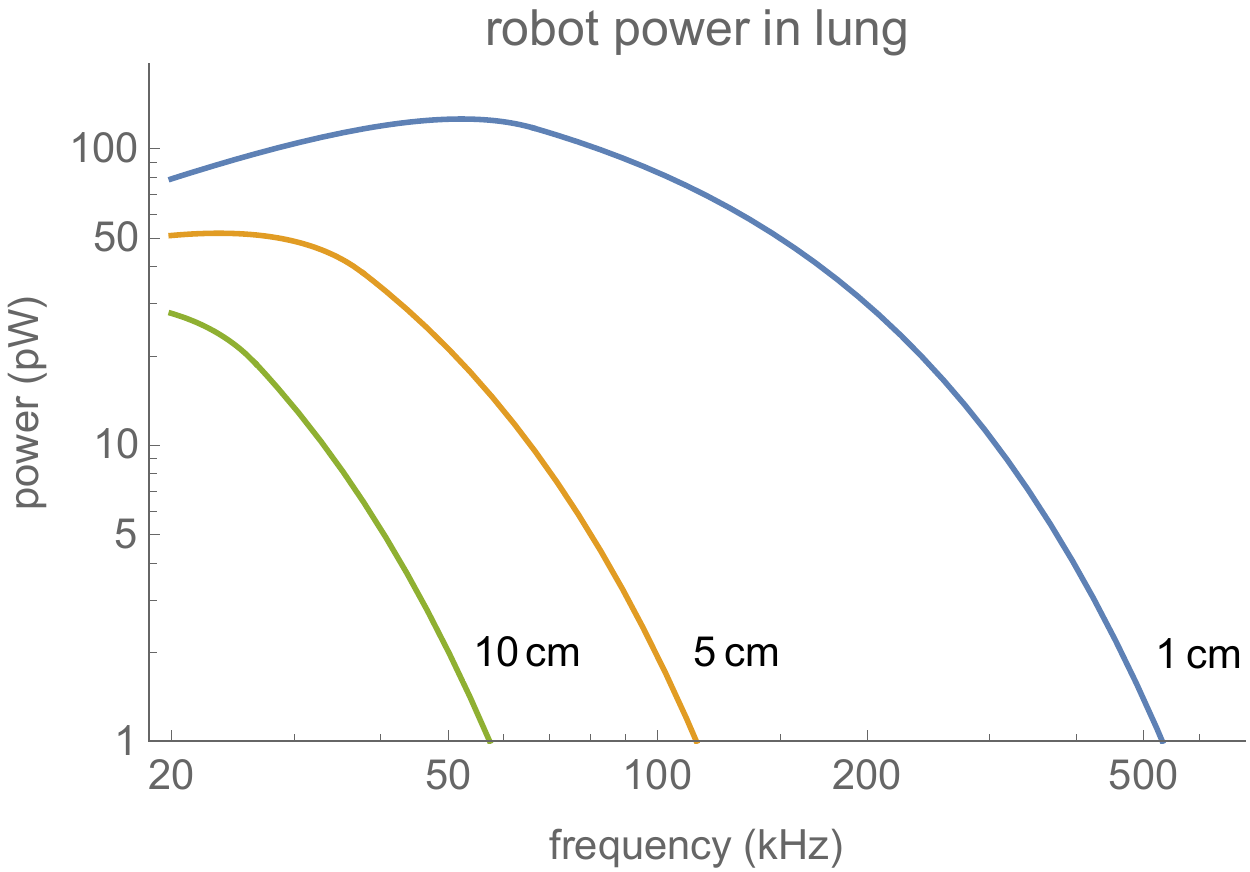} \\
\end{tabular}
\caption{Power vs.~frequency for a robot at various distances from the ultrasound transmitter in soft tissue and lung, for the scenario of \tbl{robot scenario}.
The value next to each curve indicates distance from the skin or from the skin-facing surface of a lung, for soft tissue and lung, respectively.}
\figlabel{power}
\end{figure}

Two robot placements illustrate the available power: in soft tissue and in a lung, which has large acoustic attenuation (see \tbl{tissue parameters}). The lung surface is about $5\,\centimeter$ beneath the skin, leading to some attenuation between the transducer on the skin and the surface of the lung. Furthermore, the different acoustic impedances of lung and soft tissue means only about $36\%$ of the acoustic energy reaching the lung is transmitted into it~\cite{freitas99}.
Additional attenuation arises from the ribs, which partly cover the lungs. Bone transmits only part of the sound reaching it, and the sound that travels through bone attenuates about 20 times more rapidly than attenuation in soft tissue~\cite{freitas99}. 
The power reaching various parts of the lung depends on their locations relative to the ribs and transducers on the skin~\cite{treeby19}. An approximate accounting for these factors estimates acoustic energy for the lung as first attenuating through $5\,\centimeter$ of soft tissue, after which $20\%$ of the incident energy enters the lung.

With these estimates, \fig{power} shows power available to a robot at various depths in soft tissue and in a lung.
The maximum power occurs near the largest frequency with sufficient pressure to move the pistons through their entire range of motion. At lower frequencies, the pistons move more slowly, resulting in less power. At higher frequencies, the reduced pressure due to increased attenuation results in slower motion and pistons do not move over their full range, also resulting in less power.
Thus the power-maximizing frequency arises from simultaneously maximizing the speed and range of piston motion.

The figure indicates that frequencies around $100\,\kilohertz$ are best for powering robots in regions with low attenuation between the robot and the sound source. Lower frequencies, e.g., $40\,\kilohertz$, are better for robots in a lung, but robots deep in a lung have much less power.
As a point of comparison, a human cell uses about $100\,\picowatt$, though with considerable variation among cell types and activity levels~\cite{freitas99,milo15}.

These results indicate how robot design modifications could provide more power. For instance, a robot could have more pistons, though that will reduce the volume and surface area available for other components.
Somewhat larger robots could accommodate more or larger pistons. This could be useful for robots implanted at a fixed location. 
Increasing robot size is not suitable for robots intended to move through circulatory system: such robots cannot be much larger than the 2-micron diameter considered here~\cite{freitas99}.

\section{Sound Attenuation by Robots}

Robots absorbing power decrease the intensity of the acoustic wave. \sect{available power} describes how much power a robot absorbs with each piston, i.e., \PTotal, of \eq{power}, both to overcome friction and provide power to other devices in the robot. In addition, microscopic robots are likely to be stiffer than biological tissues~\cite{freitas99}. 
This gives the robots significantly different acoustic properties than tissue, which lead to scattering and dissipation in boundary layers around each robot. This section evaluates these losses, which occur whether or not a robot absorbs energy, and compares them to attenuation due to robot energy harvesting.

A useful measure of how an object affects sound is its cross section: the ratio of scattered or absorbed power, over a cycle of the wave, to the time-average flux of the incident acoustic wave, given by \eq{flux}. This area can differ substantially from the geometric surface or cross section areas of the object.

\subsection{Uniform Surface Response to Acoustic Pressure}\sectlabel{uniform surface response}

The sound wavelengths considered here are much larger than the robot size (see \tbl{sound waves}). This simplifies the evaluation of scattering and dissipation around the robot. In particular, the long wavelength means the sound is insensitive to the distribution of features on the robot surface. Moreover, the acoustic pressure at a given time is nearly the same over the entire robot. So all pistons move in phase with each other. Thus, instead of modeling each piston individually, we can estimate a robot's effect on the sound by treating the piston motion induced by acoustic pressure variation as its average value spread uniformly over the robot's surface.

Specifically, \eq{piston motion} gives each piston's motion in response to acoustic pressure $p(t)$. The constant-force spring cancels the ambient pressure, so $dx/dt = -p(t) A/\kTotal$. Due to the robot's stiffness, the rest of the surface has negligible motion in response to the pressure. With pistons covering a fraction \fPistons\ of the surface, the corresponding uniform response is the average of these two cases, i.e., treating the entire surface as having radial velocity $-\beta p(t)$ where $\beta =\fPistons A/\kTotal$.
For the robot described in \tbl{robot scenario}, 
$\fPistons = 11\%$.
With the parameters of \tbl{piston parameters}, the full range of piston motion corresponds to about 2\% change in sphere radius from this average motion. This leads to approximating the robot as a slightly vibrating sphere with radius \rRobot.

This approximation of uniform surface motion and a plane wave impinging on the sphere gives an axisymmetric sound field. This sound produces elastic waves within the sphere. However, such waves have a minor effect for the long wavelengths considered here~\cite{faran51}, which allows neglecting elastic waves and instead using a specified boundary condition on the sphere's surface. Thus the sound around a sphere in response to an incident plane wave is approximately that arising from the boundary condition that the radial velocity of the sphere's surface is the sum of $-\beta p(t)$ and the radial component of the fluid velocity induced by the plane wave.
For long wavelengths, the scattered sound is much weaker than the incident wave~\cite{hilgenfeldt98}. This allows replacing $p(t)$ in the boundary condition by the pressure of the incident plane wave. With this simplification, the boundary condition does not depend on the scattered pressure, and evaluating the scattered sound follows the same procedure as for scattering from rigid objects and gas bubbles~\cite{fetter80,hilgenfeldt98,sullivan-silva89}, but with the boundary conditions for a moveable surface described here.

\subsection{Scattering}

For the simplified situation described in \sect{uniform surface response}, the scattering cross section in the long-wavelength limit considered here (see \tbl{sound waves}) is~\cite{fetter80,sullivan-silva89}
\begin{equation}\eqlabel{scattering cross section}
\sigma = \pi \rRobot^2 \left( 4 (k\, \rRobot)^2  (\beta c \rho)^2 + \frac{4}{9}  (k\, \rRobot)^4 \left(1-12  (\beta c \rho)^2 \right) \right)
\end{equation}
as described in \sectA{scattering cross section}.
The first factor, $\pi \rRobot^2$, is the geometric cross section of the robot. Since $k \rRobot\ll1$, the scattering cross section is much smaller than the robot's size.

A hard sphere corresponds to $\beta=0$, for which \eq{scattering cross section} becomes $(4/9)\pi \rRobot^2 (k \rRobot)^4$~\cite{hilgenfeldt98}.
This is Rayleigh scattering, proportional to the fourth power of the sound's frequency~\cite{rayleigh15}.
If the sphere's center of mass is held in place rather than allowed to move in response to the sound, the cross section is a bit larger: the fraction is $7/9$ instead of $4/9$.

\subsection{Viscous and Thermal Dissipation}\sectlabel{dissipation}

\begin{table}[tp]
\begin{center}
\begin{tabular}{lcl}
dynamic viscosity	&\viscosity	&$10^{-3}\,\pascal \; \second$	\\
bulk viscosity		&\viscosityBulk	&$3\times10^{-3}\,\pascal \; \second$	\\
temperature	&\Tfluid	&$310\,\Kelvin$ \\
thermal conductivity	&\kThermal	&$ 0.6\,\watt/\meter/\Kelvin $ \\
heat capacity	&\heatCapacity		& $ 4200\,\joule/\kilogram/\Kelvin $ \\
ratio of heat capacities	&\heatCapacityRatio		& $1.02$ \\
\end{tabular}
\end{center}
\caption{Viscous and thermal properties of the fluid around the robot, taken to be close to those of water and blood plasma~\cite{freitas99,holmes11}. The heat capacity is at constant pressure and \heatCapacityRatio\ is the ratio of the value at constant pressure to that at constant volume.}\tbllabel{fluid parameters}
\end{table}

Near the robot surface, viscous and thermal effects alter the sound propagation, which leads to dissipation. This occurs in viscous and thermal boundary layers with characteristic lengths $\sqrt{2\viscosity/(\density \omega)}$ and $\sqrt{2\kThermal/(\density \heatCapacity \omega)}$ for viscous and thermal dissipation, respectively~\cite{fetter80,fleckenstein18}.
For the scenarios considered here, 
these boundary layers extend about a micron from the robot surface. 

To evaluate these effects, we use viscous and thermal properties of water, given in \tbl{fluid parameters}. This is reasonable for robots in the bloodstream: because the boundary layers are small, the properties of blood determine behavior in the boundary layers around the robots, rather than the properties of the tissues through which the vessels are passing. This contrasts with using tissue attenuation properties for the sound propagation (see \tbl{tissue absorption coefficients}) since the wavelengths are large compared to size of most vessels.

The dissipation due to viscosity and thermal conduction in these boundary layers is proportional to $(4/3)\viscosity + \viscosityBulk$ and $\kThermal(\heatCapacityRatio-1)/\heatCapacity$, respectively.
With the parameters used here, these values are $4\times 10^{-3}\,\pascal\;\second$ and $3\times 10^{-6}\,\pascal\;\second$, respectively.
Thus, as is commonly the case for sound propagation in liquids, viscous dissipation is much larger than that due to thermal conduction.

Analytic models of sound scattering can include these boundary layers~\cite{fleckenstein18}. Alternatively, numerical evaluation of the dissipation readily incorporates the approximate uniform surface motion in response to the sound described above. This gives dissipation from both the tissue (see \tbl{tissue absorption coefficients}) and the boundary layers around the sphere. Numerically evaluating the pressure with viscous and thermal dissipation~\cite{fetter80} for cases with and without the sphere allows determining the increased dissipation due to the presence of the robot. Without the sphere, dissipation arises only from tissue attenuation. The difference between these two cases gives the dissipation due to the viscous and thermal boundary layers. As expected for sound in liquids, viscous dissipation is much larger than that due to thermal effects. Dividing this dissipation by the flux of the incident plane wave gives the cross section for dissipation in the fluid around the sphere.

Viscous losses in the boundary layer around an object arise from no-slip condition at its surface. Appropriately designed surfaces could reduce this dissipation, as discussed in \sect{piston viscous drag}.
In addition, the robot surface may be built to allow tangential motion, e.g., because combinations of both radial and tangential motion can improve the efficiency of locomotion based on surface vibrations~\cite{hogg14}.
A surface with some tangential motion, along with the radial piston motion, could more closely match the acoustic motion of fluid when there is no robot. Such a surface will have less effect on the sound wave than a tangentially rigid surface, and hence less viscous dissipation.

\subsection{Attenuation Cross Sections for a Robot}\sectlabel{single-robot attenuation}

\begin{figure}
\centering
\includegraphics[width=\figwidth]{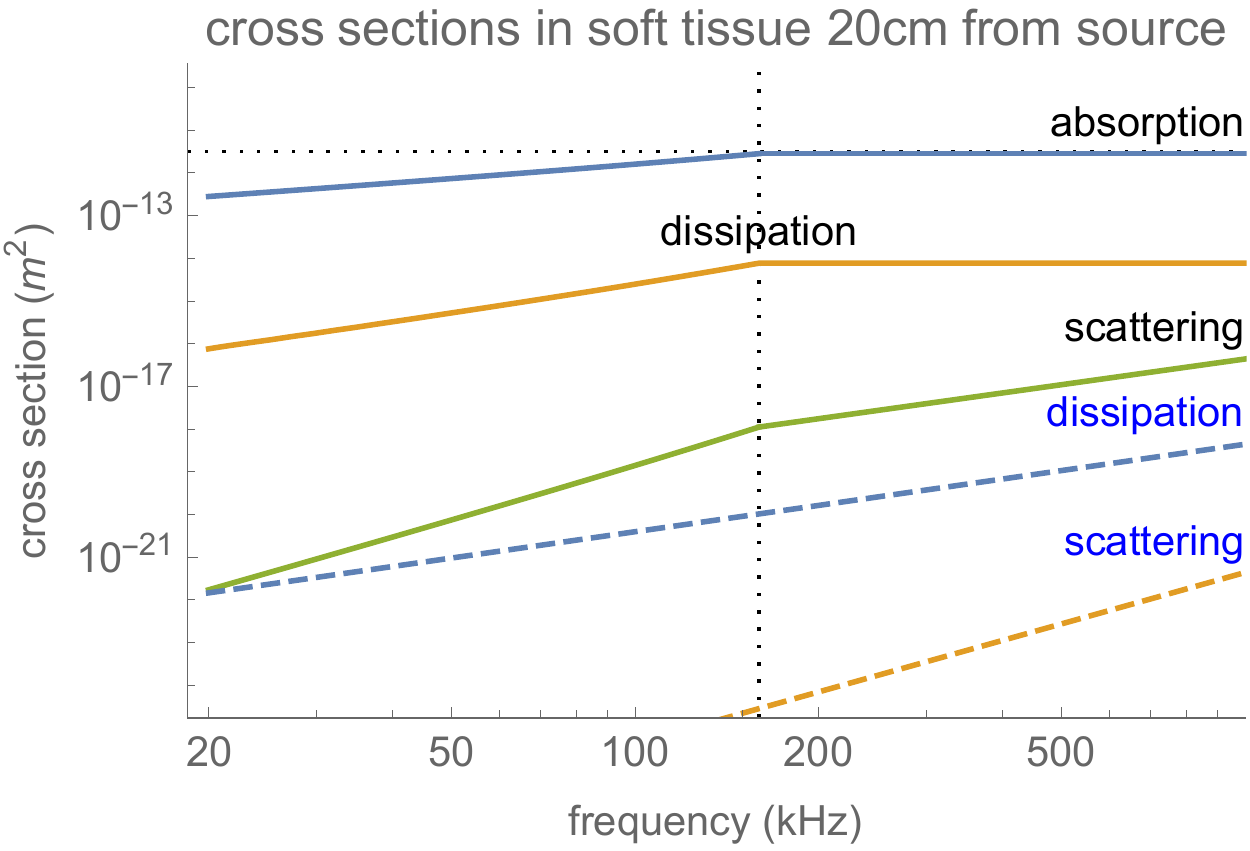}
\caption{Cross sections vs.~frequency for a single robot, with 20 pistons, located in soft tissue $20\,\centimeter$ from a $50\,\kilopascal$ transducer. The curves show cross sections for energy absorbed by the pistons, dissipation in the fluid around the robot and energy scattered by the robot.
For comparison, the dashed curves show cross sections for dissipation and scattering by a hard sphere, corresponding to a robot with pistons locked at their limit of motion so they don't respond to the acoustic pressure.
The dotted horizontal line is the geometric cross section of the robot, $\pi \rRobot^2$. The vertical dotted line is the frequency, $160\,\kilohertz$, at which $\lambda=2$ (see \eq{motion parameters}). 
}\figlabel{cross section}
\end{figure}

\fig{cross section} shows the contributions of processes removing energy from a plane wave for a robot in soft tissue $20\,\centimeter$ from the ultrasound source, the same scenario as in \fig{power}. The total cross section for removing energy from the sound wave is the sum of these three contributions. As illustrated in this figure, absorption is the dominant contribution, i.e., the pistons extract significantly more energy from the sound than dissipation in boundary layers, which in turn is much larger than the scattered energy. 
The figure also shows cross sections for a hard sphere, corresponding to locked pistons when the robot is not collecting energy. These are much less than when robots absorb power.

\subsection{Sound Attenuation by a Swarm of Robots}\sectlabel{many-robot attenuation}

\begin{table}
\centering
\begin{tabular}{ccc}
number		& typical spacing		& number density \\ \hline
$10^{10}$		& $170\,\micron$		& $2\times10^{11}/\meter^3$ \\
$10^{11}$		&  $80\,\micron$		& $2\times10^{12}/\meter^3$\\
$10^{12}$		&  $40\,\micron$		& $2\times10^{13}/\meter^3$\\
\end{tabular}
\caption{Scenarios with various numbers of robots in a body with volume $\bodyVolume = 50\,\liter$. The typical spacing is the average distance between neighboring robots, estimated as the cube root of the average volume per robot.
}\tbllabel{scenarios}
\end{table}

Prior studies of powering microscopic robots focus on one or a few nearby robots~\cite{freitas99,hogg10}. This is adequate to evaluate the power available to individual robots and how power distributes among nearby robots. In the case of acoustic power, \fig{cross section} shows the cross sections for a single robot are very small. Thus, a few microscopic robots will not significantly alter the intensity of sound with wavelengths substantially larger than the robots. 
This means that the power available to a robot is not affected by a few other robots at distances large compared to the robot's size.

However, proposed applications involve large numbers of robots~\cite{freitas99}. These robots not only compete for power directly with nearby robots, but also deplete the power source globally. This is a particular concern for acoustics, where the power source is outside the body: robots in the body could significantly attenuate sound reaching deep into the body, analogous to the increased attenuation due to bubbles~\cite{ainslie11}.

Thus accessing the applicability of acoustic power to swarms of microscopic robots requires estimating how large numbers affect sound propagation. To do so, this section applies the effect of a single robot, described in \sect{single-robot attenuation}, to evaluate attenuation due to many robots for the scenarios in \tbl{scenarios}. 

In the scenarios considered here, robots are randomly positioned in the body, so sound scattering from the robots is incoherent.
Thus attenuation from a group of robots is the sum of that from each robot individually. In this case, robots with number density $\numberDensityRobot$ and cross section $\sigma$ give amplitude attenuation
\begin{equation}\eqlabel{attenuation with robots}
\attenuationRobot = \frac{1}{2} \numberDensityRobot \sigma
\end{equation}
The attenuation for power is twice this value. 

For robots uniformly distributed throughout the body \numberDensityRobot\ is the ratio of number of robots to the body volume. There could be some variation in this value. E.g., if robots are uniformly distributed in the blood volume, then parts of the body with high or low blood supply will have correspondingly higher or lower \numberDensityRobot.  
If, instead, robots concentrate in a portion of the body, e.g., a single organ, the corresponding number density for a given number of robots will be larger in that region, with correspondingly larger attenuation, and smaller elsewhere.

\begin{figure}
\centering 
\begin{tabular}{cc}
\includegraphics[width=\mfigwidth]{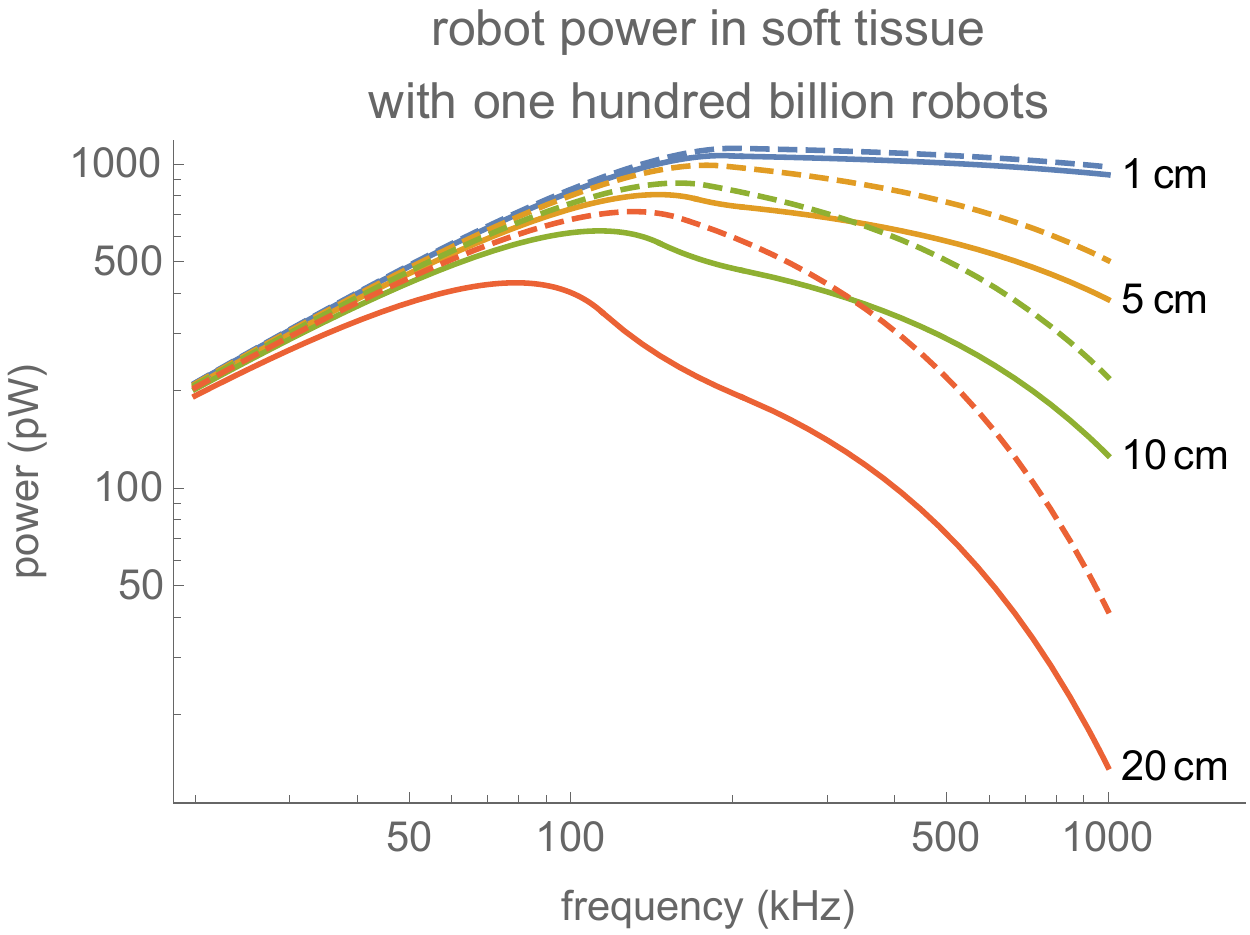} & \includegraphics[width=\mfigwidth]{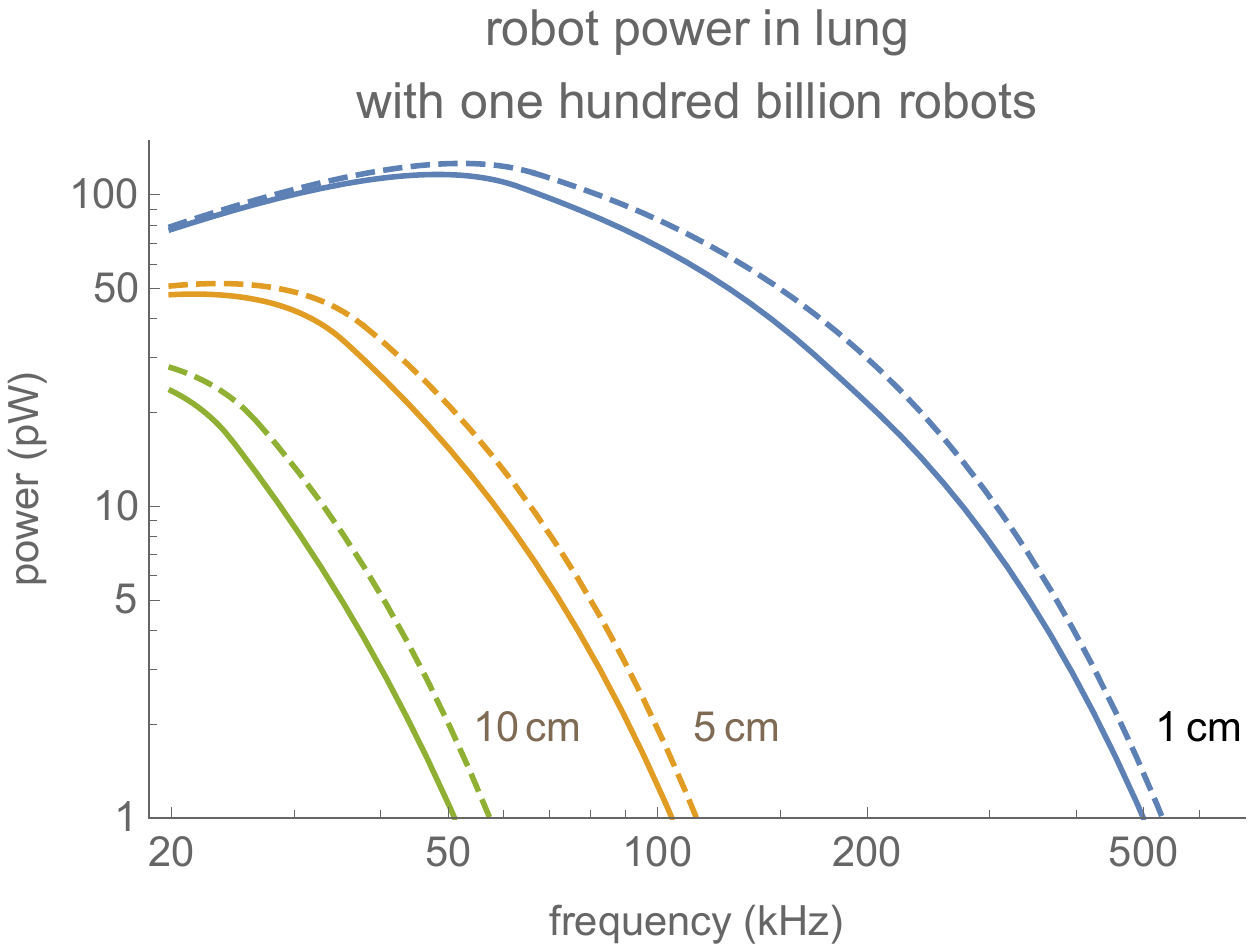} \\
\end{tabular}
\caption{Power vs.~frequency with attenuation due to $10^{11}$ uniformly-distributed robots collecting power (solid curves) for the scenario of \tbl{robot scenario}. The dashed curves show power without robot attenuation, from \fig{power}.
}\figlabel{power with robot attenuation}
\end{figure}

\begin{figure}
\centering 
\includegraphics[width=\widefigwidth]{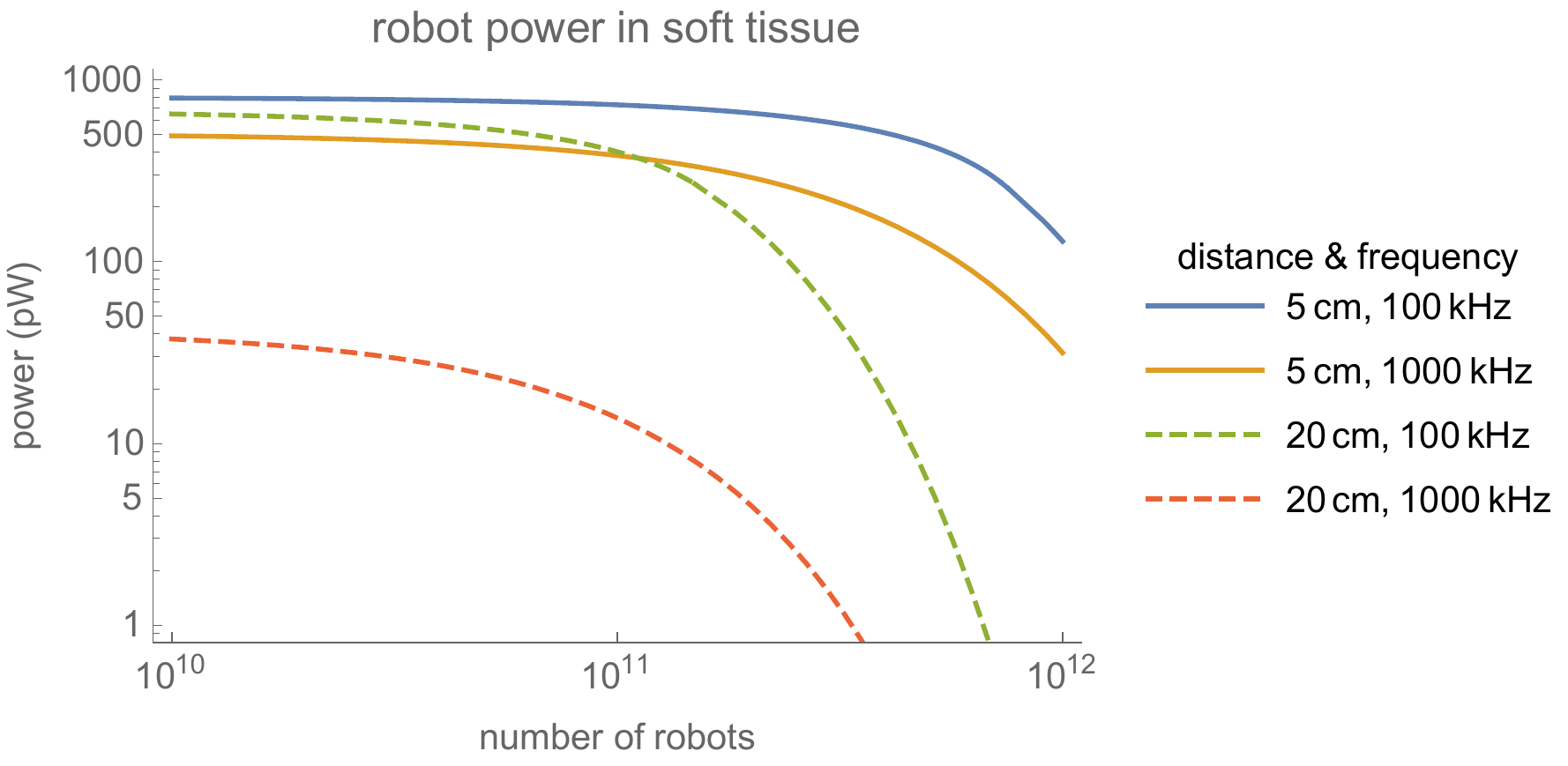}
\caption{Power with attenuation due to robots collecting power in soft tissue as a function of number of robots uniformly distributed in the body volume for the scenario of \tbl{robot scenario}. The curves correspond to the indicated distances from the source and sound frequencies.
}\figlabel{power vs number of robots}
\end{figure}

\fig{cross section} shows that energy absorbed by the pistons is the dominant contribution to a robot's acoustic cross section. Thus we use power absorption to evaluate the effect of large numbers of robots. 
As an example, \fig{power with robot attenuation} shows how $10^{11}$ robots affect available power. Comparing with \fig{power} shows the robots significantly reduce power in soft tissue at the higher frequencies and distances. On the other hand, attenuation in lung is so large that robots add a relatively minor amount to the attenuation.
Attenuation increases with additional robots and thereby results in less available power, as shown in \fig{power vs number of robots} for robots in soft tissue.

For the scenario of \fig{power with robot attenuation}, averaging over locations in the body, $10^{11}$ robots could each collect around $100\,\picowatt$, so the total power collected by all the robots is tens of watts. In this example, transducers deliver $1000\,\watt/\meter^2$ over much of the body's surface, so the total power is about $1000\,\watt$.
Thus robots extract only a small portion of the delivered acoustic power.

Robot power extraction depends on pressure (see \sect{piston power}). Hence the robots' contribution to attenuation is pressure-dependent.
At low frequencies and small distances, pressure is large, so the piston range limits the amount of energy robots can extract from the acoustic wave. In that case, robots add relatively little to the attenuation. However, when pressure decreases enough so that pistons do not move through their full range of motion, robots extract significant power and greatly increase the attenuation. Since pressure decreases with distance from the transducers, robot-induced attenuation is particularly large far from the transducers, i.e., where power is already relatively low due to tissue attenuation.

\section{Compensating for Swarm Attenuation}\sectlabel{compensating for attenuation}

\sect{many-robot attenuation} shows that a swarm of microscopic robots significantly increases attenuation in tissue.
Unlike passive particles, microscopic robots can coordinate their behavior~\cite{hogg06a} to mitigate this problem by altering the distribution of sound in both time and space to flexibly allocate power among robots in the swarm.
This section describes and evaluates several mitigation strategies robots could use individually or in combination, possibly at different times or locations.
This swarm behavior can be particularly beneficial for robot missions with temporally distinct phases requiring different amounts of power in different locations depending on each robot's local environment.

\subsection{Adjusting Mission Scope and Timing}

A direct way to avoid significant attenuation is selecting tasks for the robots that do not require large numbers, e.g., using well below $10^{12}$ in soft tissue (see \fig{power vs number of robots}). However, this may limit the speed or effectiveness of applications, particularly those requiring significant concentrations of robots acting at the same time, such as collecting data simultaneously from many cells, comparing observations with those of neighboring robots, or requiring many robots to act simultaneously on many cells.

Another strategy to cope with low power far from the source is for robots to adjust their activities based on available power. For example, robots have more power when they are near the skin than when deep in tissue. Thus a robot could defer power-intensive tasks, such as data analysis or communication, until they are near the skin. Conversely, robots deeper in tissue could perform only the most immediately relevant tasks, or only operate intermittently.

\subsection{Storing Energy}\sectlabel{stored energy}

This study evaluates the steady-state power available from acoustic pressure. Instead of using the power as received, a robot could store energy for later use~\cite{freitas99}. Stored energy can provide higher burst power, or power when the robot is in a location where the sound is too attenuated to provide adequate power.
Energy storage allows separating when energy is collected from when it is used. 
A limitation of this approach is that storage mechanisms could require a significant fraction of the robot volume.

Robots that move through the body continually change their distances from the skin. If robots have sufficient energy storage, they could collect enough energy when near the skin to power their activities deeper in the body, or when they pass into high-attenuation tissues, such as lung.
An example is robots delivering oxygen~\cite{freitas98} to supplement that from red blood cells. The main activity of such robots is pumping oxygen and carbon dioxide into or out of tanks. This occurs while they are in capillaries, which is only a few percent of the time they circulate in the blood~\cite{freitas99}.  
These robots would have little acoustic power while in the lung, so would need to collect energy prior to entering the lung.

Energy storage could also be useful when using a mix of robot sizes. That is, larger robots, stationed at fixed locations rather than being small enough to travel through capillaries, could collect and store energy. By vibrating their surfaces, such robots could act as small transducers distributed in the tissue, providing power to smaller robots as they pass nearby. This power transmission would occur over much smaller distances than power from external transducers, so could use higher frequencies that would significantly attenuate if used by the external transducers. These larger robots could exploit the acoustic efficiency of higher frequencies~\cite{hogg12} to provide both power and communication, while receiving power over longer distances by using lower frequencies.

\subsection{Limiting Robot Power Collection}

\begin{figure}
\centering 
\includegraphics[width=\figwidth]{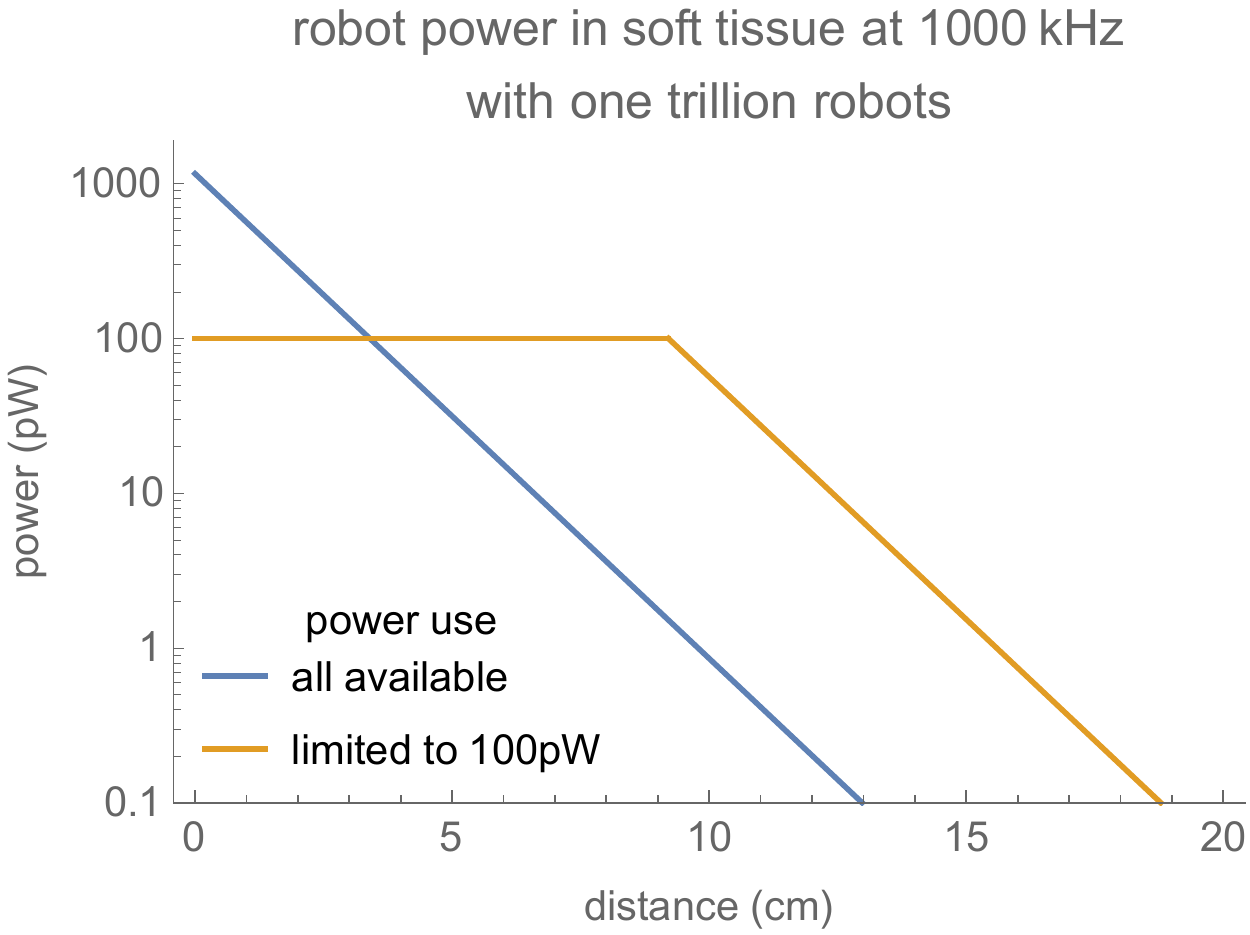}
\caption{Power with $10^{12}$ robots for sound with $1000\,\kilohertz$ for the scenario of \tbl{robot scenario}. The curves show power when robots use all available power and when they each use at most $100\,\picowatt$.
}\figlabel{limited power}
\end{figure}

Acoustic attenuation produces a highly nonuniform distribution of power in the body. Robots absorbing power increase attenuation and thus increase this variation. This means robots near the skin have a great deal of power, while those deeper in the body have little. 

Robots near the skin could reduce this power gradient by limiting their power collection, e.g., by locking some pistons in place.
Alternatively, a robot could reduce the power collected by each piston. This could involve slower motion (by increasing the load on each piston) or stopping pistons during a portion of each acoustic cycle (e.g., by reducing the range of piston motion). For the same power, the latter approach leads to more dissipation, and hence higher attenuation. Thus slowing piston speed is the better option.

Stopping piston motion over several acoustic cycles rather than just part of one has an additional benefit if robots near the skin synchronize their duty cycles for absorbing power: while all robots near the skin stop absorbing power, they reduce sound attenuation, making bursts of higher power available to deeper robots. This contrasts with unsynchronized duty cycles, which only somewhat increases average power to deeper robots. A synchronization signal could be added to the sound wave from transducers, or could be provided from clocks in the robots.

Another option is to select the range of pistons so that high pressure variation pushes them to their limits. While stuck at their limits, pistons do not absorb power.
In this case, robots less than half a wavelength apart would automatically be approximately synchronized by experiencing extreme pressure variations at about the same time. This is particularly useful for robots intended to operate near the skin. Such robots can get significant power without a large range for pistons and will have more of their volume available for other uses.

Collecting less power reduces the major contribution to acoustic attenuation even though these robots still attenuate sound through dissipation in the boundary layer and by scattering (see \fig{cross section}).
As an example of this strategy, \fig{limited power} shows a situation with such a large number of robots that little power remains $20\,\centimeter$ from the source (see \fig{power vs number of robots}). In this case, robots that limit their power considerably increase the depth at which other robots receive substantial power.

\subsection{Using Multiple Frequencies}\sectlabel{multiple}

\begin{figure}
\centering 
\includegraphics[width=\figwidth]{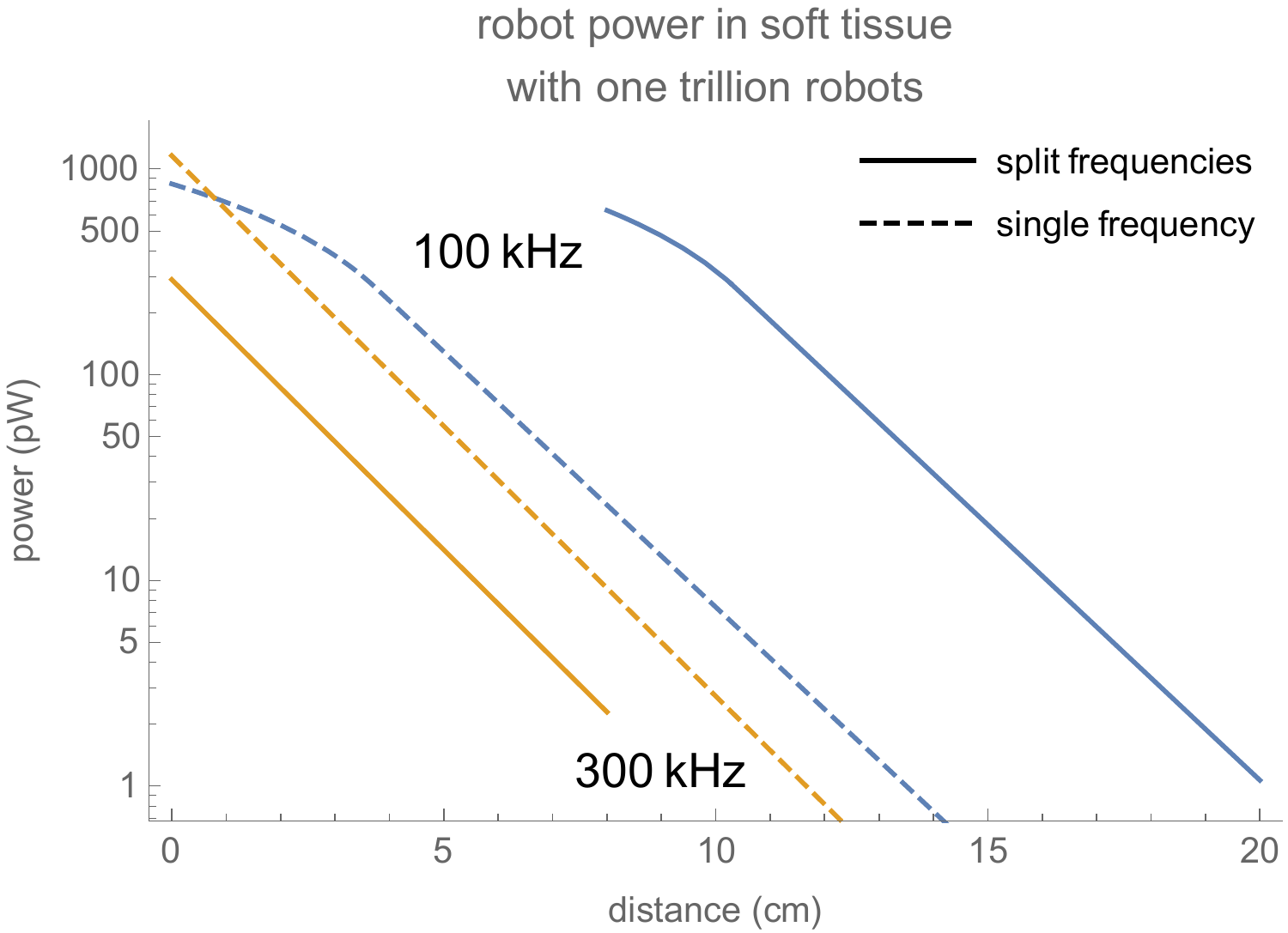}
\caption{Power with $10^{12}$ robots. The solid curves show power for the scenario of \tbl{robot scenario}, except the $1000\,\watt/\meter^2$ source intensity is split between $100\,\kilohertz$ and $300\,\kilohertz$, with source pressures $43\,\kilopascal$ and $23\,\kilopascal$, respectively. 
Robots use the higher frequency at small depths and the lower frequency at large depths. 
The dashed curves show power available when all robots extract power from just one of the frequencies, each with source intensity of $1000\,\watt/\meter^2$.
}\figlabel{split frequencies}
\end{figure}

Using two or more frequencies can provide more power to robots deeper in the body.
With this approach, robots near the skin extract power from a higher frequency, while a lower frequency, with lower attenuation, is reserved for deeper robots. By not extracting power from the low frequency, these robots only passively attenuate the sound, i.e., by scattering and viscous dissipation in fluid around the robot. This passive attenuation is significantly less than attenuation due to power extraction (see \fig{cross section}).

A limitation of this approach is that the combined intensity of the frequencies must not exceed the safe limit on total intensity. Thus splitting the sound among two or more frequencies means less power from each frequency than if using that frequency alone.

\fig{split frequencies} is an example of this strategy. Robots monitor the power available from the higher frequency, and use that frequency for power when it provides at least $2.3\,\picowatt$, which occurs at a distance of $8\,\centimeter$ from the source. Otherwise they switch to the lower frequency. 
In this case, robots $10\,\centimeter$ from the source have more power than those a bit closer to the skin, so acoustic power is not a monotonic function of distance into the body.
This split-frequency approach provides significantly more power to deeper robots than when all robots use just one of these frequencies.

To apply this technique, robots near the skin adjust their springs to avoid responding to the low frequency, in the same way they compensate for variations in ambient pressure, as discussed with \eq{piston motion}. 
As an example, adjusting the constant-force spring to compensate for $100\,\kilohertz$ pressure variation requires changing the overlap, \Lspring, much more rapidly than needed to adjust for changes in ambient pressure. As an extreme case, consider a robot close to a source with $50\,\kilopascal$ pressure amplitude. It must adjust for pressure changing by twice this amount each half-period of the acoustic wave. This corresponds to changing $\Lspring$ by about $35\,\nanometer$ in $5\,\microsecond$. From \eq{F drag for spring} and using the upper bound on $\Lspring \hspring$ of $0.01\,\micron^2$, this adjustment dissipates less than $10^{-3}\,\picowatt$. While far larger than the power dissipated to adjust to ambient pressure variation (see \sect{piston viscous drag}), this dissipation is much less than the power available to the robot.

\subsection{Positioning Robots to Provide Low-Attenuation Paths}

Robots avoiding locations between skin and a region of deeper tissue can allow more sound to reach robots in that deeper region. These regions may not be static, e.g., some tissue moves a few centimeters during each breath, which is larger than the wavelengths considered here. Robots must adjust for this motion if it changes their positions relative to the deeper locations that require the additional power. They could do so either locally, by comparing positions relative to nearby robots, or via external signals from imaging, such as used with radiological treatments~\cite{korreman15}.

If deeper robots do not require continuous power, robots could employ this mitigation only as needed, thereby alternating power between deeper and shallower robots.
This could occur on a periodic schedule to provide a predictable duty cycle of power to the deeper robots,
Alternatively, the occasional power needs of deeper robots could occur in response to events in their local environment, such as when sensing specific patterns of chemicals that require significant computational analysis or communication to other robots. In such cases, the robots could communicate their need for additional power, either directly or via neighbors, to shallower robots, requesting that those robots cease their power use to allow more power to reach the deeper robot.
This approach to occasional power is an alternative to robots collecting and storing energy for occasional burst use (see \sect{stored energy}). Creating low-attenuation paths to deeper tissue is particularly attractive when onboard energy storage would require too much of the robot volume to support the required burst power.

With this method of sharing acoustic power, robots manage energy collection and use as a group, rather than each robot focusing only on its own energy requirements.
This approach is especially useful if the main power-using activities are in relatively small deep regions of the body.

\begin{figure}
\centering 
\includegraphics[width=\widefigwidth]{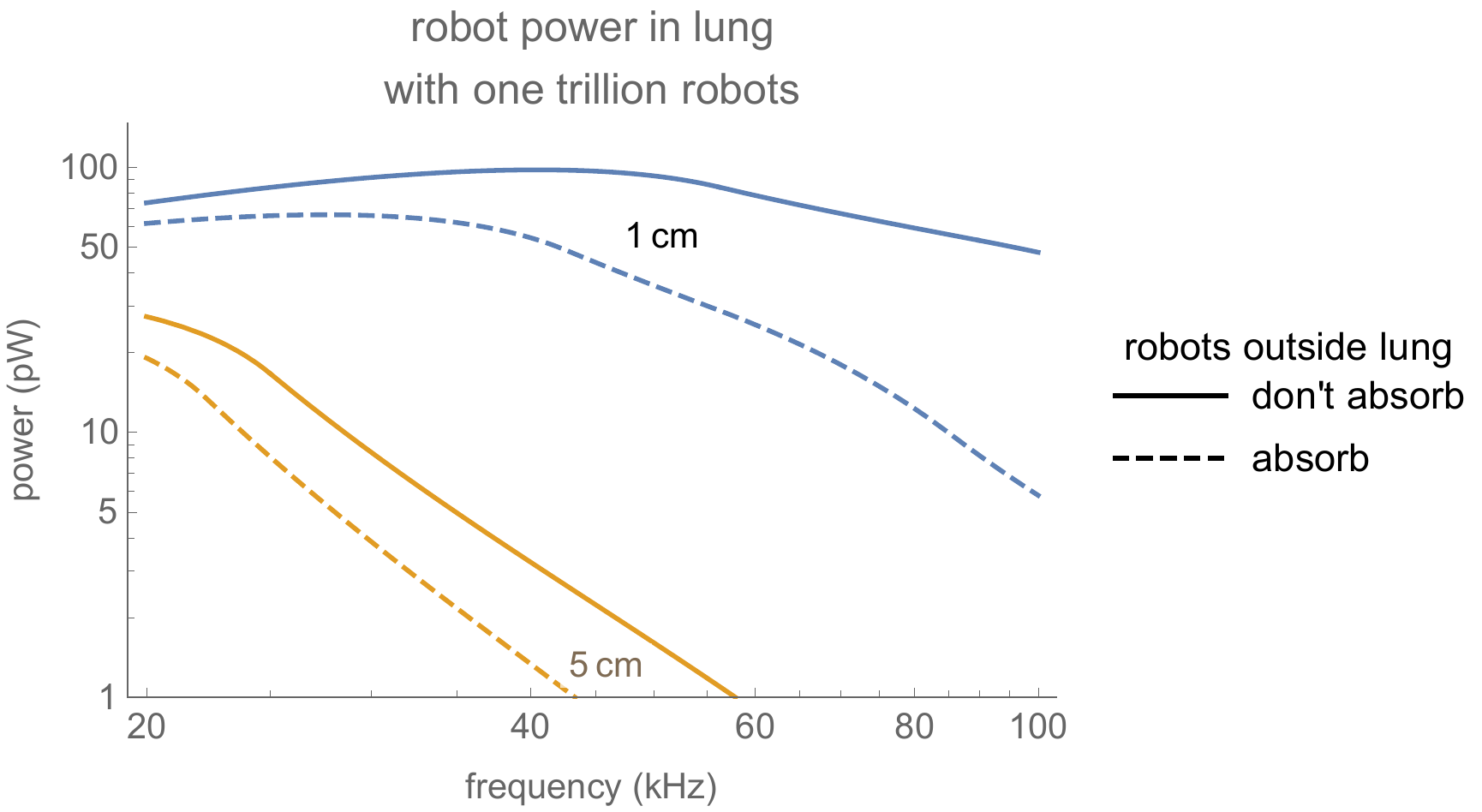}
\caption{Power vs.~frequency for robots in a lung when there are $10^{12}$ robots in the body for the scenario of \tbl{robot scenario}. The solid curves show power when robots avoid locations between the lung surface and the skin. The dashed curves show power when uniformly distributed robots absorb power at all locations, including between the lung and the skin. The values next to the curves indicate distances from the skin-facing surface of the lung.
}\figlabel{avoid path}
\end{figure}

\fig{avoid path} is an example of this strategy where robots avoid paths between transducers on the skin and the surface of the lung. In this case, the number of robots is large enough to considerably attenuate the sound in the tissue between the lung and the skin (see \fig{power vs number of robots}). Avoiding those locations provides more power to robots in the lung.

Ideally, for this strategy, the robots would actively avoid the path regions. Thus they would not contribute to attenuation by any of the mechanisms in \fig{cross section}. However, moving robots out of some regions leads to higher concentrations in others, thereby increasing attenuation in those regions above the values described in \sect{many-robot attenuation} for a uniform distribution of robots.

Even if robots cannot control their movements, they could achieve much of the same effect by stopping or reducing their power absorption when they detect they are passing through path locations, e.g., indicated by navigation signals~\cite{freitas99}. Identifying these locations would only need resolution comparable to a wavelength or larger, i.e., millimeters. Such robots would still attenuate sound due to scattering and dissipation near their surfaces, but much less than when they absorb power (see \fig{cross section}).

Generalizing from this example, methods that design swarm behaviors to form specific global shapes~\cite{werfel14,slavkov18} and with specific collective behaviors~\cite{altemose19} could be applied to produce desired patterns of acoustic waves.
Combining these swarm design techniques with how robots affect acoustic waves, as discussed in \sect{many-robot attenuation},
could  allow a swarm to adjust available power in ways suited for specific robot missions.
For example, robots able to move independently and measure their distances to neighbors could position themselves to precisely tune acoustic properties at scales well below the wavelength, thereby creating dynamic acoustic metamaterials~\cite{chen10c,kadic13,zheludev10}. In these materials, sound scatters coherently from the robots, in contrast to incoherent scattering used to evaluate the effect of many robots (see  \eq{attenuation with robots}).
An application of such materials is focusing sound into small, deeper regions where some of the robots require additional power.
Metamaterials can respond selectively to specific frequencies. 
This capability could enhance power distribution when robots use multiple frequencies, as described in \sect{multiple}. Specifically, the positioning of the swarm could direct the lower frequency toward robots that are deeper in the tissue to deliver power more precisely.

\subsubsection{Heterogeneous Robots}\sectlabel{heterogeneous}

The examples of available power in this paper use the scenario of \tbl{robot scenario}. More generally, a swarm could consist of robots with different designs, such as using different numbers or sizes of pistons. This heterogeneity can improve swarm performance when different types of robots are best suited for different tasks required for their overall mission~\cite{kengyel15}.

For example, in some applications, robots may remain in one location for an extended period of time, e.g., to collect measurements on individual cells throughout the cell cycle. This task could benefit from different types of robots. E.g., robots designed to work in high-power environments near the skin could only use high frequencies and hence could have fewer, shallower pistons, since their range of motion is less at high frequencies. Such pistons would not be as effective at collecting power from lower frequencies. On the other hand, robots intended for deeper operation could devote more of their volume to pistons to more efficiently collect power from lower frequencies.
This, a heterogeneous mixture of robots could better match the availability of acoustic power than if all robots have the same design. Moreover, evolutionary algorithms can adapt the fraction of different types in such heterogeneous swarms~\cite{kengyel15}.

Different designs could be useful even if robots continually change their distance from the skin, e.g., as they move with blood in the circulation. In this case, robots could usually operate with a small baseline power for their controller, possibly from stored energy. Then, when they find themselves in an environment suited to their design, they could activate their full operation. Due to the large number of robots in the swarm, this intermittent activation can still have many activated robots at all times.

\section{Conclusion}

\begin{figure}
\centering
\includegraphics[width = \widefigwidth]{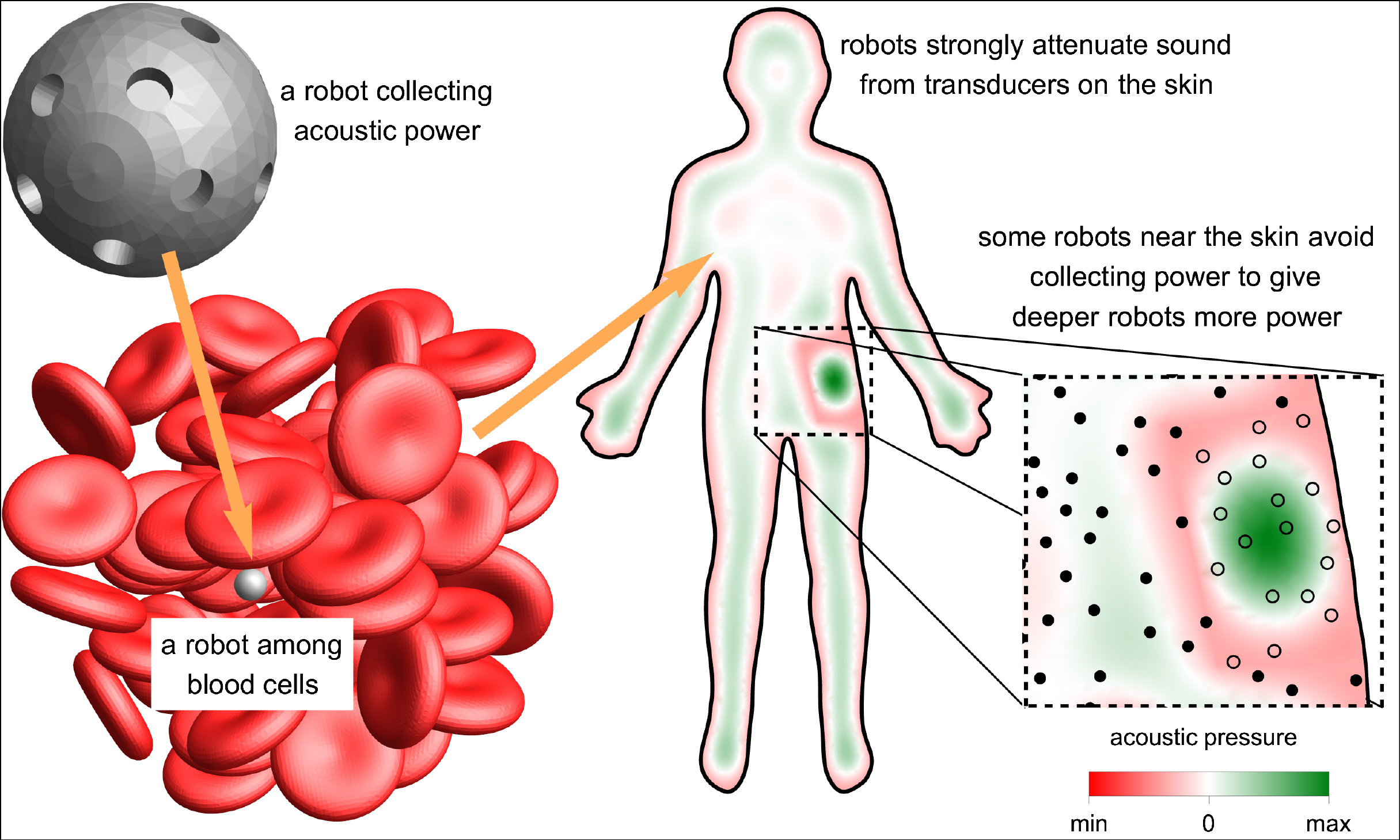}
\caption{Summary of a swarm of microscopic robots arranging for power to reach robots deeper in the body. From left to right, a single robot with pistons collects power from acoustic waves, a robot among blood cells, schematic of sound highly attenuated in the body due to absorption by many robots, and a region of robots near the skin avoiding power collection (open circles) to allow deeper robots (solid points) to receive more energy. The robots considered here are smaller and more numerous than indicated in this schematic diagram of robots in tissue.}
\figlabel{summary}
\end{figure}

This paper evaluated ultrasonic power for swarms of microscopic robots using mechanical energy harvesters. 
Frequencies around $100\,\kilohertz$ provide the most power in much of the body. 
However, robots at a depth of more than a centimeter or so in high-attenuation tissue receive little power. Robots in such locations require other power sources~\cite{freitas99}, or they must store enough energy to support their operation.
Even within a tissue with low attenuation, power can vary with small-scale changes in robot location. This is because power decreases with increasing viscosity of the fluid around the robot (see \eq{available power}). The scenarios of this paper correspond to robots in blood vessels. Robots outside of blood vessels could encounter materials with much larger viscosities~\cite{freitas99}, and hence have less power than nearby robots in vessels. 

When the number of robots in the body exceeds about a hundred billion (corresponding to a total mass of a few tenths of a gram), the robots significantly increase acoustic attenuation, thereby reducing the power available to robots deep in the body. Similar attenuation can occur locally, e.g., in a single organ, with fewer robots that concentrate at those locations.
This reduction would not be apparent in experiments involving a small number of robots. Thus a method to power implanted devices that is adequate for a few robots could fail when applied to large numbers. 
In such situations, designing missions for microscopic robots must account for their collective effects. Robots with relatively modest sensing, communication and computation capabilities can mitigate this attenuation with swarm behaviors such as those discussed in \sect{compensating for attenuation} and illustrated in \fig{summary}. These behaviors can flexibly deliver power to robots where and when that is most important for their medical task.

Providing acoustic power to microscopic robots throughout the body requires continually operating transducers over a substantial fraction of the body surface so all robots are within a relatively short distance of a transducer. 
This raises two challenges.

First, treatments using transducers over much of the body would only occur in clinical settings. This is not suitable for long-term applications such as ongoing health monitoring. Instead, such acoustic power for long-term use would be delivered by only one or a few small transducers a person could continually wear~\cite{wang22}. The power of such transducers would be limited by battery storage, and only provide significant power to small regions of the body near those transducers.

A second challenge arises from the total power applied by the transducers.
While $1000\,\watt/\meter^2$ is safely used routinely for imaging  over a small area of the body~\cite{freitas99}, such transducers covering a significant fraction of the body surface, 
would deliver about $1000\,\watt$. This compares with the body's basal power use of about $100\,\watt$.
For applications lasting a significant time, e.g., hours, this level of heating could require active cooling. If this is not feasible, long-lasting applications would require reducing sound intensity, thereby reducing the power available to the robots. Such reduction could be continuous, or occur through pauses in operation. These pauses could be brief, e.g., a duty cycle that powers the transducers one millisecond out of every 10. Such pauses would be tolerable if robots can store enough energy to continue operation between acoustic bursts, or can accomplish their tasks with intermittent operation, e.g., for diagnostics that only need to sample the robot's environment after it moves to a new location.

Acoustic power poses challenges for robot design. While pistons are relatively simple energy collectors, they are also bulky, using a significant fraction of robot volume and surface, compared, for example, to chemical power from fuel cells~\cite{freitas99,hogg10}. Thus using enough pistons to collect adequate power may reduce performance of other uses of the robot volume or surface.

For example, a robot may require substantial volume for tanks of chemicals, energy storage or information processing.
The robot could require surface area for a variety of tasks.
One important task is surface devoted to chemical sensors or pumps. Such devices can be much smaller than the total surface area, allowing room for a large number of them~\cite{freitas99}. For this task, pistons are not likely to be a major issue because chemical collection is only weakly dependent on the fraction of surface area absorbing the chemical~\cite{berg93}. 
However, area devoted to pistons could be more limiting for other uses.
For example, communication~\cite{hogg12} or locomotion~\cite{hogg14} can require substantial surface area. A possible mitigation to these competing uses is for the pistons to perform multiple functions. For example, the actuators adjusting the spring's force in response to changing ambient pressure could also provide force to move the pistons. This could readily provide surface oscillations for locomotion, which involve lower frequencies ($1\mbox{--}10\,\kilohertz$) but similar range of motion~\cite{hogg14} as the pistons considered here. 
Frequencies comparable to those used to receive power may be useful for robots to selectively destroy some types of diseased cells next to the robot~\cite{mittelstein20}.
On the other hand, actuating the pistons for acoustic communication requires much higher frequencies and smaller ranges of motion~\cite{hogg12}, which may be beyond the capabilities of actuators involved in power collection by adjusting the springs in response to ambient pressure changes.

In light of these limitations, both for the patient and robot design, the relative simplicity of acoustic power is best suited to applications where robots perform relatively limited tasks that don't require lengthly on-going operation or many additional mechanisms inside, or on the surface of, the robot. These situations are particularly relevant for early applications of microscopic robots.
More broadly, these considerations of the feasibility and challenges of acoustic power show that the choice of how to power microscopic robots strongly depends on the nature of their intended application and the number of robots used.

The results of this paper can help identify applications of swarms of microscopic robots where acoustic power is a good option. Since the total collected power is the product of the number of pistons on the robot and the power collected by each one, the power evaluated here for spherical robots also applies to other shapes that contain the same pistons, providing flexibility in selecting the robot shape. These results can inform the design of such robots when they become feasible to manufacture in large numbers. Conversely, the mitigating swarm behaviors discussed here can reduce the complexity, of both hardware and software, that each robot would need to operate in spite of the enhanced attenuation from other robots, e.g., by including sufficient onboard energy storage and associated control computations, in the absence of such mitigation. This could somewhat simplify the formidable challenge of manufacturing the microscopic robots considered here.

There are several directions for future study of acoustic power for swarms of microscopic robots. 
One direction is designing and building acoustic energy harvesters that best exploit the limited robot volume and surface area with the constraint of viscous drag on motion in the fluid around the robot. Reducing the space required for powering the robot will leave more room for other components, thereby extending the range of missions robots can perform. One such improvement is developing nanoscale engineered surfaces with low viscous drag, as discussed in \sect{piston viscous drag}.
These harvesters must not only effectively extract energy from acoustic waves, but also ensure biocompatibility. For example, this requires minimizing the rate at which surrounding materials stick to the moving surfaces and inhibit their operation. Moreover, the immune reaction to small devices can depend on their shape as well as surface chemistry~\cite{yasa20}. Thus biocompatibility requirements for the duration of the robots' mission may constrain design choices for the number, size, range of motion and shape of the pistons.

Another direction is evaluating how large numbers of robots affect structured acoustic fields used to move robots~\cite{mohanty20,rao15,shaglwf19}, which are more complex than the long-wavelength plane waves considered here. 
Thus it will be important to evaluate changes to the field structure, and forces they can apply, in addition to overall attenuation of the field strength.
If these robots also absorb energy for their internal use, using different frequencies for internal robot power and the structured fields would reduce competition among these uses of the sound. Nevertheless, such uses would still have dependencies due to safety limits on total acoustic intensity delivered to the body, similar to that seen for the dual-frequency approach shown in \fig{split frequencies}.

A third direction of study is more precisely evaluating available power in the complex acoustic environment of the body, including the effect of heterogeneous tissues, with varying acoustic properties~\cite{treeby19}.
For example, the acoustic analysis described here focuses on longitudinal compression waves. This is reasonable for evaluating the sound absorbed by robots in the bloodstream. In general, propagation could be affected by other properties of biological tissues. In particular solid structures such as bone can also support shear waves, and wave propagation can vary with direction, i.e., forming nonisotropic elastic solids instead of simple fluids. These material properties could distribute acoustic energy differently than would be the case for purely longitudinal waves.
Robot mission designs will need to account for these variations, which depend on both a robot's macroscale position in the body and the robot's immediate environment, e.g., within tens of microns. 
Available power will also depend on the locations and activities of other robots, both nearby and between the robot and the acoustic sources on the skin. Thus it will be important to evaluate how the mitigating strategies described in this paper perform with the variation in tissue properties and how swarms can adapt to those variations to support a wide variety of high-precision medical applications.

\section*{Acknowledgements}

I have benefited from discussions with Robert Freitas~Jr., Ralph Merkle, Matthew Moses and James Ryley.

\appendix

\section{Appendix: Scattering Cross Section}\sectlabel{scattering cross section}

We evaluate the sound scattered by a sphere of radius $a$ in response to an incident plane wave using the boundary conditions described in \sect{uniform surface response}: the radial velocity of the sphere's surface equals the sum of $-\beta \pIncident(t)$ and the radial component of the periodic fluid velocity induced by the plane wave on the sphere's center of mass. For evaluating scattering, we neglect dissipative effects, which are treated separately in \sect{dissipation}.

For this discussion, we express pressure and velocity in terms of complex-valued amplitudes $p$ and $v$, respectively, and time-dependence $e^{-i \omega t}$ where $\omega = 2\pi f$ is the angular frequency. The actual values are the real parts, e.g., the pressure is $\Re(p e^{-i \omega t})$, where $\Re$ denotes the real part. For acoustic waves, pressure and velocity amplitudes are related by~\cite{fetter80} 
\begin{equation}\eqlabel{velocity amplitude}
v = -\frac{i}{\omega \density} \nabla p
\end{equation}
and the amplitudes satisfy the Helmholtz equation
\begin{equation}
\nabla^2 p + k^2 p = 0
\end{equation}
where $k=\omega/\soundSpeed$ is the wave number.
The time-average flux is
\begin{equation}\eqlabel{flux from amplitudes}
\flux = \frac{1}{2} \Re(v p^*)
\end{equation}
where $p^*$ is the complex-conjugate of $p$~\cite{fetter80}.

A convenient representation for matching boundary conditions on a sphere is expressing the pressure in spherical coordinates centered on the sphere with the $z$ direction corresponding to the incident plane wave's motion. The total sound pressure is the sum of incident and scattered waves: $p = \pIncident + \pScattered$.

The pressure amplitude of the incident plane wave propagating along the $z$-axis is $\pIncident = p_0 e^{i k z} $ where $p_0$ is the pressure magnitude of the wave. 
In spherical coordinates, $z = r \cos(\theta)$ so expressing the plane wave in solutions to the wave equation in spherical coordinates gives the pressure as a sum over modes~\cite{fetter80,sullivan-silva89}:
\begin{equation}
\pIncident = p_0 \sum_{m=0}^\infty F_m P_m(\cos \theta) j_m(k r)
\end{equation}
where $F_m = i^m (2m+1)$, $P_m$ is the Legendre polynomial of order $m$, and $j_m$ is the spherical Bessel function of order $m$.
From \eq{velocity amplitude}, the corresponding velocity amplitude, in the $z$ direction, is $\vIncident = p_0/(\density \soundSpeed) e^{i k z}$. 
The radial component of this velocity is $\vIncident \cos(\theta)$.

Similarly, the amplitude of the scattered pressure,  \pScattered, is an outgoing wave with expansion~\cite{fetter80}
\begin{equation}
\pScattered = p_0 \sum_{m=0}^\infty A_m P_m(\cos \theta) h_m^{(1)}(k r)
\end{equation}
where $h_m^{(1)}$ is the spherical Hankel function of the first kind of order $m$, and the coefficients $A_m$ are determined by matching the boundary condition at the surface of the sphere, as described below.

At the surface of the sphere, the radial component of the sound's velocity amplitude matches the specified boundary condition. From \eq{velocity amplitude}, this condition is
\begin{equation}
 -\frac{i}{\omega \density} \frac{\partial p}{\partial r} = -\beta \pIncident + \vIncident \cos(\theta)
\end{equation}
evaluated at $r=a$, the radius of the sphere and \vIncident\ evaluated at the sphere's center of mass, i.e., $z=0$.
Substituting the above expansions for the incident and scattered waves in this equation gives a relation that must hold for all polar angles $\theta$. This requires matching each mode separately, thereby determining the coefficients $A_m$ of the scattered wave.
In particular, since $P_1(x)=x$, the incident velocity \vIncident\ only contributes to mode $m=1$.

The time-average flux of the incident plane wave, \fluxIncident, from \eq{flux from amplitudes} equals \eq{flux} with $p$ replaced by $p_0$. \eq{flux from amplitudes} gives the scattered flux, \fluxScattered, in terms of the coefficients $A_m$. For a spherical surface $S$ of radius $r$ centered on the scattering sphere, the total scattered power is the integral of \fluxScattered\ over the surface $S$. Thus the scattering cross section is
\begin{equation}\eqlabel{scattering cross section def}
\sigma = \frac{1}{ \fluxIncident} \int_S \fluxScattered dS
\end{equation}
where $dS = r^2 \sin(\theta) d\theta d\phi$ is the differential surface area for the sphere, with polar and azimuthal angles, $\theta$ and $\phi$, respectively. The scattered sound spreads over the sphere but is not attenuated in this model. Thus the integral is independent of radius $r$.

When the sound wavelength is large compared to the sphere's radius, i.e., $k a \ll 1$, only modes $m=0$ and $m=1$ contribute significantly to the scattered sound. This long-wavelength limit applies to the scenarios considered here (see \tbl{sound waves}). In this case, the coefficients are
\begin{equation}\eqlabel{coefficients}
\begin{split}
A_0 	&= \frac{2}{3} \beta  c \density (k a)^4 -\frac{1}{3} i (k a)^3 - \beta  c \density (k a)^2\\
A_1	&= -\frac{1}{2} i \beta  c \density (k a)^4
\end{split}
\end{equation}
Using these values in \eq{scattering cross section def} gives the scattering cross section of \eq{scattering cross section} for the sphere with radius $a=\rRobot$.

\end{document}